\documentclass[10pt,twocolumn,letterpaper]{article}

\usepackage{cvpr}
\usepackage{times}
\usepackage{epsfig}
\usepackage{graphicx}
\usepackage{amsmath}
\usepackage{amssymb}
\usepackage{mathtools, nccmath}

\usepackage{booktabs}  

\usepackage{bm}
\usepackage{microtype}
\usepackage{amsmath}
\usepackage{amssymb}
\usepackage{amsthm}
\usepackage{array}
\usepackage{multirow}
\usepackage{cases}
\usepackage{slashbox}

\usepackage{algorithm}
\usepackage{algpseudocode}
\newtheorem{thm}{Theorem}
\newtheorem*{thm1}{Theorem}
\newtheorem*{thm2}{Theorem}

\theoremstyle{remark}

\newcommand{\beginsupplement}{%
	\setcounter{table}{0}
	\renewcommand{\thetable}{S\arabic{table}}%
	\setcounter{figure}{0}
	\renewcommand{\thefigure}{S\arabic{figure}}%
}

\usepackage[pagebackref=true,breaklinks=true,letterpaper=true,colorlinks,bookmarks=false]{hyperref}

\cvprfinalcopy 

\def\cvprPaperID{1891} 

\ifcvprfinal\fi
\begin{document}

\title{Sliced Wasserstein Generative Models}
\author{Jiqing Wu*$^{1}$ \quad
Zhiwu Huang*$^{1}$ \quad
Dinesh Acharya$^{1}$ \quad
Wen Li$^{1}$ \quad \\
Janine Thoma$^{1}$ \quad 
Danda Pani Paudel$^{1}$ \quad
Luc Van Gool$^{1,2}$\\
$^{1}$ Computer Vision Lab, ETH Zurich, Switzerland \quad $^{2}$ VISICS, KU Leuven, Belgium\\
{\tt\small \{jwu, zhiwu.huang, liwen, jthoma, paudel, vangool\}@vision.ee.ethz.ch}\\
{ *\tt\small Equal contribution }
}

\maketitle

\begin{abstract}
In generative modeling, the Wasserstein distance (WD) has emerged as a useful metric to measure the discrepancy between generated and real data distributions. Unfortunately, it is challenging to approximate the WD of high-dimensional distributions. In contrast, the sliced Wasserstein distance (SWD) factorizes high-dimensional distributions into their multiple one-dimensional marginal distributions and is thus easier to approximate.

In this paper, we introduce novel approximations of the primal and dual SWD. Instead of using a large number of random projections, as it is done by conventional SWD approximation methods, we propose to approximate SWDs with a small number of parameterized orthogonal projections in an end-to-end deep learning fashion. As concrete applications of our SWD approximations, we design two types of differentiable SWD blocks to equip modern generative frameworks---Auto-Encoders (AE) and Generative Adversarial Networks (GAN).

In the experiments, we not only show the superiority of the proposed generative models on standard image synthesis benchmarks, but also demonstrate the state-of-the-art performance on challenging high resolution image and video generation in an unsupervised manner \footnote{Code: \url{https://github.com/musikisomorphie/swd.git}}.

\end{abstract}

\section{Introduction}
\label{submission}
The Wasserstein distance (WD) is an important metric, which was originally applied in the optimal transport problem\footnote{Optimal transport addresses the problem of finding an optimal plan to transfer a source distribution to a target distribution at minimal cost.}~\cite{villani2008optimal}. 
Recently,~\cite{arjovsky2017wasserstein,gulrajani2017improved,salimans2018improving,wei2018improving,tolstikhin2017wasserstein,liu2018two,adler2018banach,ambrogioni2018wasserstein} discovered the advantages of the WD in generative models and achieved state-of-the-art performance for image synthesis.
However, the WD has some drawbacks. For instance, its primal form is generally intractable for high-dimensional probability distributions, although some works~\cite{tolstikhin2017wasserstein,Genevay2017learning,salimans2018improving} have proposed relaxed versions of the primal form.
The dual form of the WD can be more easily derived, but it remains difficult to approximate its $k$-Lipschitz constraint~\cite{wei2018improving,wu2018wass}.

Given the weaknesses of the WD, the sliced Wasserstein distance (SWD) suggests itself as a potential alternative. 
The SWD factorizes high-dimensional distributions into their multiple one-dimensional marginal distributions and can therefore be approximated more easily, as studied in~\cite{bonneel2015sliced,kolouri2016sliced,kolouri2017optimal,deshpande2018generative}.
However, due to the inefficient approximations of the SWD in existing methods, its potential in generative modeling has not been fully explored yet.

In this paper, we address this issue. Our contributions can be summarized as follows:
\begin{itemize}     
    \item Relying on our novel primal SWD approximation, we propose sliced Wasserstein Auto-Encoder (SWAE) models.
    By seamlessly stacking the proposed primal SWD blocks (layers) on top of the standard encoder, 
    we give the traditional AE generative capabilities.
    State-of-the-art AE-based generative models usually require an additional regularizer to achieve the same effect.
    \item Based on our new dual SWD approximation, we introduce a sliced version of Wasserstein Generative Adversarial Networks (SWGAN) by applying the proposed dual SWD blocks to the discriminator. To the best of our knowledge, this is the first work to study the dual SWD and its application to generative models.
    \item To satisfy the orthogonality constraint required by the projection matrices of the proposed SWD blocks, we apply a non-Euclidean optimization algorithm on Stiefel manifolds to update the projection matrices. 
    \item Motivated by the improvements of visual quality and model stability demonstrated on the standard image synthesis benchmarks, we apply our proposed model to the challenging task of unsupervised high resolution image and video synthesis. These evaluations confirm the advantage of our model under non-trivial cases.
\end{itemize}

\section{Background}
\subsection{Wasserstein Distance \& Related Models}

The primal Wasserstein distance (WD) is given by
	\begin{equation}
	W_p(P_X, P_Y) = \inf_{\gamma \in \Pi(P_{X}, P_{Y})} \mathbb{E}_{(X, Y)\sim \gamma}  [d^p(X, Y)]^\frac{1}{p},
	\label{eq:primal}
	\end{equation}
where $X, Y$ are random variables, $\Pi(P_{X}, P_{Y})$ denotes the set of all joint distributions $\gamma(X, Y)$ whose marginal distributions are $P_{X}, P_{Y}$ respectively, $d$ is a metric, and $p > 0$.
For $p = 1$, the Kantorovich's dual of the WD is
\begin{equation}
\label{eq:dual}
W_1(P_X, P_Y) = \sup_{f \in \mathrm{Lip}^1} 
\mathbb{E}_{X\sim P_{X}}[f(X)] - \mathbb{E}_{Y\sim P_{Y}}[f(Y)],
\end{equation}
where $ \mathrm{Lip}^1$ is the set of all $1$-Lipschitz functions. 
The dual WD becomes $k \cdot W_1$ if we replace $ \mathrm{Lip}^1$ with $ \mathrm{Lip}^k$ for $k > 0$.

The original form of the primal WD (Eq.~\ref{eq:primal}) is generally intractable.
For the case of Auto-Encoders (AE), however,~\cite{tolstikhin2017wasserstein} have proven that optimizing the primal WD over tractable encoders is equivalent to optimizing it over the intractable joint distributions $\gamma(X, Y)$.
This idea yields Wasserstein Auto-Encoders (WAE). 
Another way of avoiding the intractable primal WD is to use its dual form instead.  
By parameterizing the $f$ in Eq.~\ref{eq:dual} with a neural network, \cite{arjovsky2017wasserstein} have found a natural way of introducing the dual WD to the GAN framework.
WAE and WGAN stand for typical applications of the primal and dual WD in generative models and are closely related to our proposed methods. We therefore summarize the necessary details in the following. 

\subsubsection{Wasserstein AE (WAE)}
The WAE proposed by~\cite{tolstikhin2017wasserstein} optimizes a relaxed version of the primal WD.
In order to impose the prior distribution on the encoder, an additional divergence $\mathcal{D}$ is introduced to the objective:
\begin{equation}
\begin{aligned}
\inf_{P_{Q(Z|X)} \in \mathcal{Q}} & \mathbb{E}_{X \sim P_X}\mathbb{E}_{Q \sim P_{Q(Z|X)}} [c(X, G(Z))]
 + \lambda  \mathcal{D}(P_Q, P_Z),
\end{aligned}
\label{Eq2}
\end{equation}
where $Z$ is random noise, $G$ is the decoder, $\mathcal{Q}$ is any nonparametric set of marginal distributions $P_Q$ on encoders $Q$, $c$ is the Euclidean distance, $\lambda > 0$ is a hyperparameter, and $\mathcal{D}$ is the divergence between the $P_Q$ of the encoder and the prior distribution $P_Z$ of $Z$.
\cite{tolstikhin2017wasserstein} instantiate $\mathcal{D}$ by using either maximum mean discrepancy (MMD) or GAN, both of which can be regarded as a distribution matching strategy.

\subsubsection{Wasserstein GAN (WGAN)}
The key challenge of WGAN is the $k$-Lipschitz constraint required in Eq.~\ref{eq:dual}.
The original WGAN~\cite{arjovsky2017wasserstein} adopt a weight clipping strategy; however, it satisfies the $k$-Lipschitz constraint poorly. To alleviate this problem, the improved training of Wasserstein GAN (WGAN-GP)~\cite{gulrajani2017improved} penalizes the norm of the discriminator's gradient with respect to a few input samples. This gradient penalty is then added to the basic WGAN loss (i.e.\ the dual form of the WD) resulting in the following full objective:
	\begin{equation}
	\begin{aligned}
	\min_{G} \max_{D} & \, \mathbb{E}_{X \sim P_X} [D(X)]  -\mathbb{E}_{\tilde{X} \sim P_G} [D(G(Z))] + 
	\\ & \lambda \, \mathbb{E}_{\hat{X} \sim P_{\hat{X}}} [(\|\nabla_{\hat{X}} D(\hat{X})\|_2-1)^2],
	\end{aligned}
	\label{Eq3}
	\end{equation}
where $G, D$ denotes the generator and discriminator respectively, $Z$ is random noise, $\hat{X}$ is random samples following the distribution $P_{\hat{X}}$ which is sampled uniformly along straight lines between pairs of points sampled from $P_X$ and $P_G$, and $\nabla_{\hat{X}} D(\hat{X})$ is the gradient with respect to $\hat{X}$.
As studied in~\cite{wei2018improving}, a limited number of samples is not sufficient to impose the $k$-Lipschitz constraint on a high-dimensional domain.
Thus,~\cite{wei2018improving} further improved the Wasserstein GAN with an additional consistency term~(CTGAN). Furthermore, \cite{miyato2018spectral} introduced a spectral normalization (SN) technique which also improves the training of GAN including the family of WGAN. The SNGAN imposes the $1$-Lipschitz constraint by normalizing the weights of each layer. 
To strengthen GAN training stability and to achieve high resolution image generation,
~\cite{karras2017progressive} applied WGAN-GP to a progressive growing scheme (PG-WGAN).

\subsection{Sliced Wasserstein Distance \& Related Models}
The idea underlying the sliced Wasserstein distance (SWD) is to decompose the challenging estimation of a high-dimensional distribution into the simpler estimation of multiple one-dimensional distributions. Formally,  let $P_{X}, P_{Y}$ be probability distributions of random variables $X, Y$. For a unit vector ${\bm{\theta}} \in \mathbb{S}^{n-1}$ we define the corresponding inner product $\pi_{\bm{\theta}}(\bm{x}) = \bm{\theta}^{T} \bm{x}$
and marginal distribution
$\pi_{\bm{\theta}}^{*} P_{X} = P_{X} \circ \pi_{\bm{\theta}}^{ -1}$.
Then the primal SWD is given by
\begin{equation}
	SW_p(P_X, P_Y) = \left(\int_{\mathbb{S}^{n-1}} W_p(\pi_{\bm{\theta}}^{*} P_{X}, \pi_{\bm{\theta}}^{*} P_{Y})^p d \bm{\theta} \right)^{\frac{1}{p}}.
\label{eq:SWD}
\end{equation}
Several works~\cite{bonneel2015sliced,kolouri2016sliced,kolouri2017optimal} exploit the fact that the WD has a closed form solution for the optimal transport plan between one-dimensional probability distributions.
More concretely, let $F_X, F_Y$ be the cumulative distribution functions (CDFs) corresponding to $P_X, P_Y$, then for all ${\bm{\theta}} \in \mathbb{S}^{n-1}$
there exists a unique closed form solution
\begin{equation}
\tau_{\bm{\theta}} = (\pi_{\bm{\theta}}^{*}F_Y)^{-1} \circ \pi_{\bm{\theta}}^{*}F_X, 
\label{eq:cummap}
\end{equation}
such that the integrand of Eq.~\ref{eq:SWD} can be computed by
\begin{equation}
	W_p(\pi_{\bm{\theta}}^{*} P_{X}, \pi_{\bm{\theta}}^{*} P_{Y})^p = \int_{\mathbb{R}} d^p(x, \tau_{\bm{\theta}}(x)) d \pi_{\bm{\theta}}^{*} P_{X} .
	\label{eq:pwd}
\end{equation}
Furthermore, as proven by~\cite{bonnotte2013unidimensional} (Chapter 5), the SWD is not only a valid distance, but also equivalent\footnote{Here, we adopt the usage of `equivalent' from~\cite{bonnotte2013unidimensional}, which is an abuse of notation.} to the WD 
\begin{equation}
SW_p(P_X, P_Y)^p \leq \alpha_1 W_p(P_X, P_Y)^p \leq \alpha_2 SW_p(P_X, P_Y)^{\frac{1}{n + 1}},
\label{eq:equ}
\end{equation}
where $\alpha_1, \alpha_2$ are constants and $n$ is the dimension of sample vectors from $X,Y$.
Given such favorable properties, the SWD has the potential to improve modern generative modeling especially when processing samples of high-dimensional distributions such as images and videos.

The SWD is typically approximated by using a summation over the projections along random directions (random projections)~\cite{rabin2011wasserstein,kolouri2016sliced,karras2017progressive,deshpande2018generative}.
For example,~\cite{rabin2011wasserstein} iteratively use a large number of random projections to estimate the SWD from samples and update the samples by gradient descent.
Similarly, the sliced Wasserstein Generator (SWG)~\cite{deshpande2018generative} optimizes its generator with an SWD loss. This SWD loss computes the difference between marginal distributions of feature maps decomposed by random projections. 
Unfortunately, these methods require a large amount of random projections and have not fully unlocked the potential of the SWD yet.

\section{Proposed Method}
Compared to a set of random vectors (projections), a set of orthogonal projections is more efficient to span an entire space.
Also, neural networks have been shown to possess robust generalization abilities.
We thus propose to approximate the primal and dual SWD with a small set of parameterized orthogonal matrices in a deep learning fashion.
In the following, we give a detailed description of our primal and dual SWD approximations. 
Later, we introduce two generative modeling applications of the resulting SWD blocks---sliced Wasserstein AE (SWAE) and sliced Wasserstein GAN (SWGAN). 

\subsection{Primal SWD Approximation}
Given $i \in \mathbb{N}$, guided by the target distribution $P_Y$, we define the $i$-th computational block which transfers the input distribution $P_{X^i}$ to $P_{X^{i+1}}$ as follows
\begin{equation}
\bm{Q}_{\bm{\Theta}}^i(\bm{x}) = \bm{O}_{\bm{\Theta}}^i \bm{\Pi}_{\bm{\Theta}}^i ((\bm{O}_{\bm{\Theta}}^{i})^T \bm{x}),
\label{eq:idtstep}
\end{equation}
where $\bm{x}$ is a sample vector from $P_{X^i}$,  $\bm{O}_{\bm{\Theta}}^i=[\bm{\theta}_1^i, \ldots, \bm{\theta}_n^i] \in \mathbb{R}^{n \times n}$ is a random orthogonal matrix, and $\bm{\Pi}_{\bm{\Theta}}^i = (\tau_{\bm{\theta}_1}^i, \tau_{\bm{\theta}_2}^i, \ldots, \tau_{\bm{\theta}_n}^i)$ (Eq.~\ref{eq:cummap}) are optimal transport maps with respect to the marginal distributions of $P_{X^i}, P_Y$ projected by $\bm{O}_{\bm{\Theta}}^i$. 
Stacking $m$ computational blocks $\bm{Q}_{\bm{\Theta}}^m \circ \ldots \circ \bm{Q}_{\bm{\Theta}}^2 \circ \bm{Q}_{\bm{\Theta}}^1$ results in the iterative distribution transfer (IDT) method~\cite{pitie2007automated}. As studied in~\cite{bonnotte2013unidimensional}, let the target distribution $P_Y$ be Gaussian, then $P_{X^m}$ converges to $P_Y$ with respect to the primal SWD, and the convergence holds when $m \rightarrow \infty$.
To reduce the number of computational blocks (Eq.~\ref{eq:idtstep}) required by IDT, we propose to parameterize the orthogonal matrices and learn them in an end-to-end deep learning fashion. 
As a result, a small number of such parameterized computational blocks is sufficient to approximate the primal SWD.

\subsection{Dual SWD Approximation}
Since the integrand of the SWD (Eq.~\ref{eq:SWD}) is nothing but a one-dimensional WD,
its Kantorovich's dual can be seamlessly applied and Eq.~\ref{eq:SWD} can be rewritten as 
\begin{equation}
 	\int_{\mathbb{S}^{n-1}} 
 	\left(\sup_{f \in \mathrm{Lip}^k} \mathbb{E}_{X_{\bm{\theta}} \sim \pi_{\bm{\theta}}^{*} P_{X}} [f(X_{\bm{\theta}})] - \mathbb{E}_{Y_{\bm{\theta}} \sim \pi_{\bm{\theta}}^{*} P_{Y}} [f(Y_{\bm{\theta}})]\right)
 	d \bm{\theta}.
\label{eq:dualSWD}    
\end{equation}
Similar to the primal SWD approximation, we propose to employ orthogonal matrices to estimate the integral over $\mathbb{S}^{n-1}$. These orthogonal matrices are also parametrized and learned in the context of deep learning. It therefore suffices to use a moderate number of orthogonal matrices to achieve a good estimation.
The proposed SWD approximations are conceptually illustrated in Fig.~\ref{fig:sotblock}.

\begin{figure}
\centering
\begin{tabular}{c}
\includegraphics[width=0.95\linewidth]{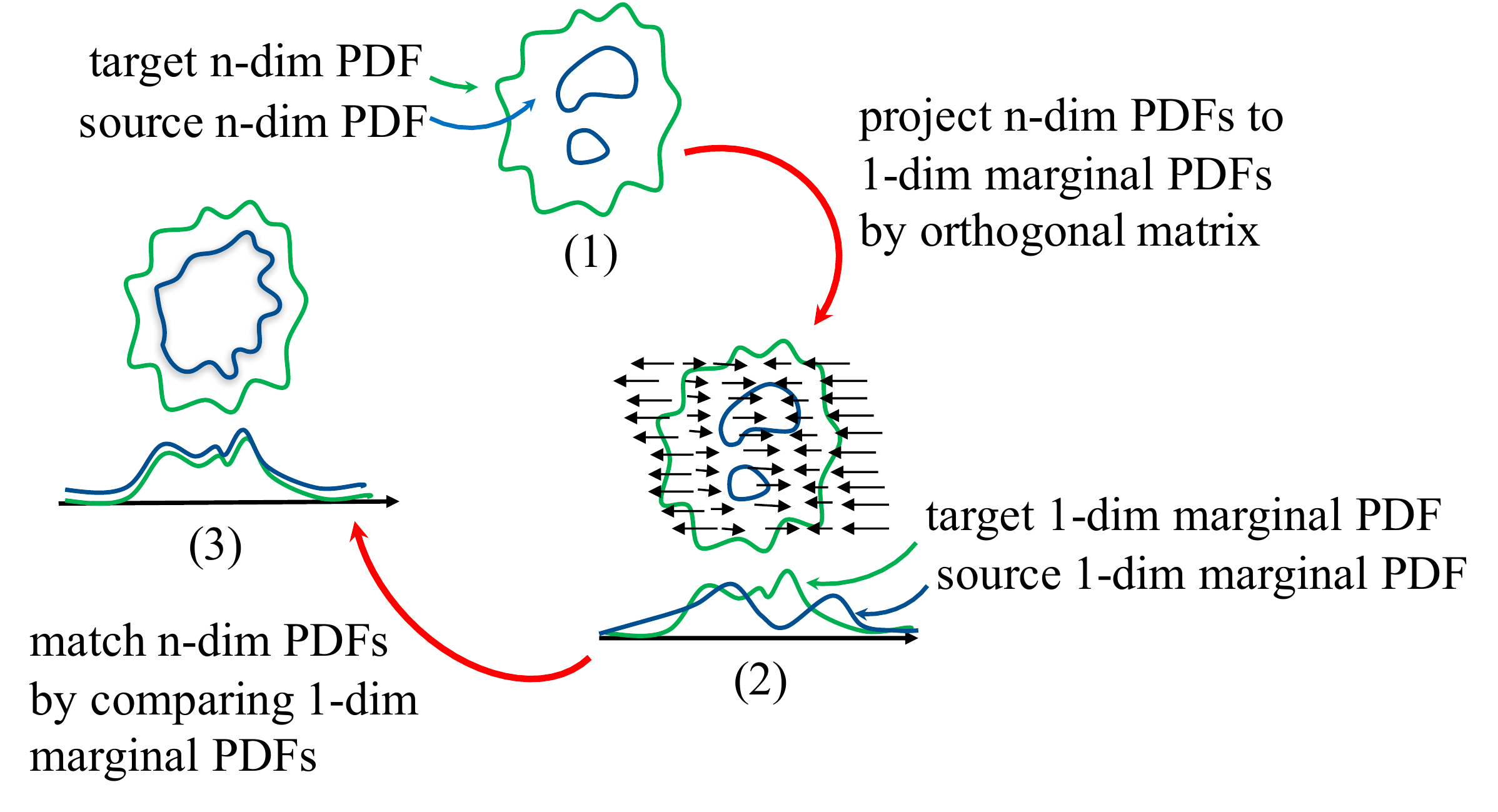}
\end{tabular}
\vspace{-0.2cm}
\caption{Illustration of our primal and dual SWD approximations. (1) - (2): By projecting samples along orthogonal unit vectors (orthogonal matrix), we decompose $n$-dimensional target and source probability distribution functions (PDFs) into their one-dimensional marginal PDFs.  (2) - (3): We match the $n$-dimensional PDFs by comparing their marginal PDFs. For the primal SWD approximation, this is done implicitly through the iterative transformation of source to target distribution. For the dual approximation, the dual SWD is calculated explicitly.}
\label{fig:sotblock}
 \vspace{-0.5cm}
\end{figure}

\begin{figure}[tbh!]
\centering
\begin{minipage}[t]{\linewidth}
\begin{algorithm}[H]
  \small
    \caption{The proposed primal SWD block}
    \label{alg:SOT}
\begin{algorithmic}
  \State {\bfseries Require:} Orthogonal matrix $\bm{O}_{\bm{\Theta}}=[\bm{\theta}_1, \ldots, \bm{\theta}_r] \in \mathbb{R}^{r \times r}$, batch of latent codes  $\bm{M}_{\bm{y}} = [\bm{y}_1, \ldots, \bm{y}_b] \in \mathbb{R}^{r \times b}$, batch of Gaussian noise $\bm{M}_{\bm{z}} = [\bm{z}_1, \ldots, \bm{z}_b]\in \mathbb{R}^{r \times b}$, and bin number $l$
  \State {\bfseries Output:} Batch of transferred latent codes  $M_{\tilde{\bm{y}}} = [\tilde{\bm{y}}_1, \ldots, \tilde{\bm{y}}_b]$
  \For{$i\gets 1, r$}
  \State $\bm{y}'_i  = \bm{\theta}_i^T \bm{M}_{\bm{y}}, \bm{z}'_i = \bm{\theta}_i^T \bm{M}_{\bm{z}}$
  \State 
  $\bm{y}''_i  = \frac{\bm{y}'_i - \min_j \{ y'_{i,j} \}} { \max_j \{ y'_{i,j} \} - \min_j \{ y'_{i,j} \}}, \bm{z}''_i  = \frac{\bm{z}'_i - \min_j \{ z'_{i,j} \}}{ \max_j \{ z'_{i,j} \} - \min_j \{ z'_{i,j} \} }$, $y'_{i,j}, z'_{i,j}$ are the $j$-th element of $\bm{y}'_i, \bm{z}'_i$ respectively.
  \State Compute soft PDF histogram $p_{y''_i}, p_{z''_i}$ of $\bm{y}''_i, \bm{z}''_i$ with $l$ bins 
  \State Compute CDF $F_{y''_i}, F_{z''_i}$ of $p_{y''_i}, p_{z''_i}$
  \State Compute $F_{y''_i}(\bm{y}''_i)$ element-wise by linear
  interpolation
  \State $ \medmath{\hat{\bm{y}}_{i} = (\max_j \{ z'_{i,j} \} - \min_j \{ z'_{i,j} \}) (F_{z''_i})^{-1} F_{y''_i}(\bm{y}''_i)  +  \min_j \{ z'_{i,j} \}}$ 
  \EndFor
    \State  Compute $\bm{M}_{\tilde{\bm{y}}} = \bm{O}_{\bm{\Theta}} \bm{M}_{\hat{\bm{y}}}^T$,     \; $\bm{M}_{\hat{\bm{y}}} = [\hat{\bm{y}}_{1}^T, \ldots, \hat{\bm{y}}_{r}^T]$
\end{algorithmic}
  \end{algorithm}
\vspace{-0.8cm}
 \begin{algorithm}[H]
 \small
    \caption{The proposed SWAE}
    \label{alg:swae}
\begin{algorithmic}
  \State {\bfseries Require:} 
  Primal SWD block number $m$, batch size $b$, decoder $G$ and encoder $Q = S_{p,m} \circ \ldots \circ S_{p,2} \circ S_{p, 1} \circ E$, 
  training steps $h$, training hyperparameters, etc.
  \For{$t\gets 1, h$}
  \State Sample real data $\bm{M}_{\bm{x}} = [\bm{x}_1, \ldots, \bm{x}_b]$ from $P_{X}$
  \State Sample Gaussian noise $\bm{M}_{\bm{z}} = [\bm{z}_1, \ldots, \bm{z}_b]$ from $\mathcal{N}(0,1)$
  \State Update the weights $\bm{w}$ of $Q$ and $G$ by descending:
  
  $\bm{w} \leftarrow \mathrm{Adam}(\nabla_{\bm{w}}( \frac{1}{b}\|\bm{M}_{\bm{x}}-G(Q(\bm{M}_{\bm{x}}, \bm{M}_{\bm{z}}))\|^2_2), \bm{w})$
  \EndFor
\end{algorithmic}
  \end{algorithm}
\end{minipage}
\vspace{-0.5cm}
\end{figure}

\subsection{Sliced Wasserstein AE (SWAE)}
Since AE-based generative models require to impose a prior distribution on the encoder, it is natural to make Eq.~\ref{eq:idtstep} learnable and incorporate it into the encoder. 
By stacking our primal SWD blocks (layers) on top of the standard encoder, we give generative capability to the traditional Auto-encoder. In other words, we can implicitly match the encoder and prior distributions without introducing an extra regularizer such as $\mathcal{D}$ required in Eq.~\ref{Eq2}.
Specifically, our encoder $Q$ is the composition of a standard encoding network $E$ and $m$ primal SWD blocks $S_{p,1}, \ldots, S_{p,m}$, that is, $Q = S_{p,m} \circ \ldots \circ  S_{p, 1} \circ E$.
By feeding the latent codes from $E$ into the primal SWD blocks $S_{p,1}, \ldots, S_{p,m}$, the distribution of latent codes is transferred to the prior distribution.
In this paper, we choose the prior distribution to be Gaussian, as it is frequently done for AE-based models. However, supported by~\cite{bonnotte2013unidimensional}, more complicated prior distributions are acceptable as well.

The implementation details of our primal SWD block are presented in Alg.~\ref{alg:SOT}.
The idea behind the algorithm is to decompose Eq.~\ref{eq:cummap} into multiple differentiable computational steps.
However, the conventional histogram computation is not differentiable.
We therefore propose a soft version of histogram computation to make the PDF histogram computation differentiable. 
More specifically, for an element $y$ we assign the weight 
$e^{-\alpha \|y - c_i\|^2}/\sum_{j=1}^l  e^{-\alpha \|y - c_{j}\|^2}$ to the $i$-th bin, where 
$c_1,\ldots, c_l$ are the bin centers. Eventually, we obtain the histogram by summing the weights for each bin over all elements $y$ . 
Note that for $\alpha \rightarrow \infty$ this soft version returns to the original non-differentiable version.
In practice, due to the minor impact of $\alpha$ on the generative capability, we empirically determine $\alpha = 1$.
As a result, the primal SWD block (Alg.~\ref{alg:SOT}) is differentiable and can be trained in a deep learning manner.

The approximation error of Alg.~\ref{alg:SOT} is dominated by its core steps corresponding to Eq.~\ref{eq:cummap}.
Since Alg.~\ref{alg:SOT} rescales all sample vectors to $[0,1]$, the inverse functions of its CDFs are again CDFs. 
Together with the fact that a CDF can be written as an empirical distribution function (EDF)~\cite{castro2015empirical}, we obtain the following error estimation for Alg.~\ref{alg:SOT}: 
\begin{thm}
\label{thm:theorem1}
Given $b \in \mathbb{N}$, let $Z_1, Z_2, \dots , Z_b $ be real-valued i.i.d.\ random variables with a continuous CDF $F_Z^{-1}$ with domain $[0,1]$.  Then we define the associated EDF $F^{-1}_{Z,b}(t) = \frac{1}{b} \sum_{i=1}^b \mathbf{1}_{\{Z_i \leq t\}}$.
Assume $\tilde{F}_Y, F_Y$ are CDFs satisfying $\|\tilde{F}_Y - F_Y \|_{\infty} \leq \gamma$, 
then there exists a $\delta > 0 $ such that for all $\varepsilon- \delta \gamma \geq\sqrt{\tfrac{1}{2b}\ln2}$ it holds that
\begin{equation}
\Pr\Bigl(\| F^{-1}_{Z,b}  \tilde{F}_Y (t) - F^{-1}_{Z}  F_Y (t)\|_{\infty} > \varepsilon \Bigr) \le e^{-2b(\varepsilon- \delta \gamma)^2}. 
\end{equation} 
\end{thm}
For a proof of Theorem \ref{thm:theorem1} please refer to our supplementary material. 
Since it is straightforward to estimate an EDF on one-dimensional data using a moderate number of samples, Theorem \ref{thm:theorem1} tells us that the core steps of our primal SWD block approximate Eq.~\ref{eq:cummap} well. 
Owing to the implicit SWD approximation by the primal SWD blocks, it is unnecessary to introduce an explicit regularization on the final objective. The objective of our proposed SWAE model is:
\begin{equation}
\inf_{P_{Q(Z|X)} \in \mathcal{Q}} \mathbb{E}_{X \sim P_X}\mathbb{E}_{Q \sim P_{Q(Z|X)}} [\|X-G(Q(X, Z))\|^2_2],
\label{eq:swae}
\end{equation}
where $Q, G$ are the encoder and decoder respectively, and $Q$ is implicitly constrained by our primal SWD blocks. The corresponding algorithm is presented in Alg.~\ref{alg:swae}.

\begin{figure}[h!]
\centering
\resizebox{\linewidth}{!}{%
\begin{minipage}[t]{9cm}
\begin{algorithm}[H]
  \small
    \caption{The proposed dual SWD block}
    \label{alg:dualSOT}
\begin{algorithmic}
  \State {\bfseries Require:} Orthogonal matrix $\bm{O}_{\bm{\Theta}}=[\bm{\theta}_1, \ldots, \bm{\theta}_r] \in \mathbb{R}^{r \times r}$ and batch of latent codes  $\bm{M}_{\bm{y}} = [\bm{y}_1, \ldots, \bm{y}_b] \in \mathbb{R}^{r \times b}$.
  \State {\bfseries Output:} Batch of $\tilde{\bm{y}}$ for dual SWD 
     \For{$i\gets 1, r$}
    \State Compute $\bm{y}'_i  = \bm{\theta}_i^T \bm{M}_{\bm{y}}$ 
    \State Compute $\bm{y}''_i = F_i(\bm{y}'_i)$ element-wise, where $F = (F_1, \ldots, F_{r})$ are one-dimensional functions to approximate the $f$ in Eq.~\ref{eq:dualSWD}.

  \EndFor

  \State $\tilde{\bm{y}} = [\bm{y}''_1, \ldots, \bm{y}''_b]^T$
\end{algorithmic}
  \end{algorithm}
\vspace{-0.8cm}
 \begin{algorithm}[H]
  \small
    \caption{The proposed SWGAN}
    \label{alg:swgan}
\begin{algorithmic}
  \State {\bfseries Require:} Number of dual SWD blocks $m$, batch size $b$, generator $G$ and discriminator $D = [S_{d,1} \circ E, \ldots , S_{d,m} \circ E ]^T$, latent code dimension $r$,
  Lipschitz constant $k$, training steps $h$, training hyperparameters, etc.
  \For{$t\gets 1, h$}
  \State Sample real data $\bm{M}_{\bm{x}} = [\bm{x}_1, \ldots, \bm{x}_b]$ from $P_{X}$
  \State Sample Gaussian noise $\bm{M}_{\bm{z}} = [\bm{z}_1, \ldots, \bm{z}_b]$ from $\mathcal{N}(0,1)$
  \State Sample two vectors $\bm{\mu_1}, \bm{\mu_2}$ from uniform distribution $U[0, 1]$ and for $l=1, \ldots, b$ calculate the elements of $\bm{M}_{\bm{\hat{x}}}, \bm{M}_{\bm{\hat{y}}}$:
  \State $\bm{\hat{x}}_l = (1- \mu_{1,l}) \bm{x}_l + \mu_{1,l} G(\bm{z}_l) $ 
  \State $\bm{\hat{y}}_l = (1- \mu_{2,l}) E(\bm{x}_l) + \mu_{2,l} E(G(\bm{z}_l)) $ 
  \State Update the weights $\bm{w}_G$ of $G$ by descending: 
 
           $\bm{w}_G \leftarrow \mathrm{Adam}(\nabla_{\bm{w}_G}(\frac{1}{b} \sum_{j,i = 1} ^{r\times m,b} D_{ji}(G(\bm{M}_{\bm{z}}))), \bm{w}_G)$
    \State Update the weights $\bm{w}_D$ of $D$ by descending:  
    
            $\bm{w}_D \leftarrow \mathrm{Adam}(\nabla_{\bm{w}_D}(\frac{1}{b} \sum_{j,i = 1} ^{r\times m,b}  (D_{ji}(\bm{M}_{\bm{x}})  -  D_{ji}(G(\bm{M}_{\bm{z}}) ) 
            +  \lambda_1 \|\nabla_{\bm{M}_{\bm{\hat{x}}}} D(\bm{M}_{\bm{\hat{x}}})\|_2 ^2 + \lambda_2 \|\nabla_{\bm{M}_{\bm{\hat{y}}}}F(\bm{M}_{\bm{\hat{y}}}) - k \cdot \mathbf{1} \|_2 ^2),\bm{w}_D)$, where we compute the gradients of $F$ element-wise.

  \EndFor
\end{algorithmic}
  \end{algorithm}
  \end{minipage}
  }
\vspace{-0.5cm}
\end{figure}

\subsection{Sliced Wasserstein GAN (SWGAN)}
The success of WGAN indicates that the dual WD can be used as a suitable objective for the discriminator of GAN models. 
In order to keep the advantages of this setup, but to avoid imposing the $k$-Lipschitz constraint on a high dimensional distribution,
we propose to use the dual SWD instead. 
Specifically, we introduce $m$ dual SWD blocks $S_{d,1}, \ldots, S_{d,m}$ to the discriminator $D$ (see Alg.~\ref{alg:dualSOT}).
Image data distributions are supported by low-dimensional manifolds. 
For this reason, classic GAN discriminators encode their input data into lower-dimensional feature maps. 
We follow this setting. 
Our discriminator is the composition of an encoding network $E$ and dual SWD blocks $S_{d,s}$, that is, $D = [S_{d,1} \circ E, \ldots, S_{d,m} \circ E]^T$. 

Eventually, we estimate the integral over $\mathbb{S}^{n-1}$ of the dual SWD  by summing over the outputs' mean value of SWD blocks $S_{d,1}, \ldots, S_{d,m}$ (see Alg.~\ref{alg:swgan}).
In order to approximate the one-dimensional optimal $f \in \mathrm{Lip}^k$ (Eq.~\ref{eq:dualSWD}) in our SWD blocks,
it suffices to use non-linear neural network layers. This is supported by the universal approximation theorem~\cite{hornik1991approximation,haykin2004comprehensive}. 
For our case, we empirically set $F_i$ in Alg.~\ref{alg:dualSOT} to be $F_i(\bm{y}'_i) =  u_{i} \mathrm{LeakyReLU}(w_{i} \bm{y}'_i + v_{i})$, where $u_{i}, v_{i}, w_{i}$ are scalar parameters.

The $k$-Lipschitz gradient penalty has its drawbacks in high dimensional space.
For one-dimensional functions, however, it can easily impose the $k$-Lipschitz constraint.
Thus, we additionally apply the gradient penalty on each dimension of the $F_i$s' output. 
Since dual WD with different $k$-Lip constraints are equivalent to each other up to a scalar, we treat $k, k'$ as tunable hyper-parameters for both $ F, D$ and relax the search interval to $k, k' \geq 0$, this relaxation can be justified by~\cite{wu2018wass}. 
Consequently, the final objective is 
\begin{equation}
	\begin{aligned}
	& \medmath{\min_{G} \max_{D} \; \int_{\theta \in \mathbb{S}^{n-1}} \Bigl(\mathbb{E}_{X \sim P_X} [D(X)]  -\mathbb{E}_{\tilde{X} \sim P_G} [D(G(Z))] \Bigr) +}
	\\ &  \medmath{\lambda_1 \mathbb{E}_{\hat{X}
	\sim P_{\hat{X}}}
	[\|\nabla_{\hat{X}} D(\hat{X}) - k'\|_2^2] + \lambda_2  \mathbb{E}_{\hat{Y}
	\sim P_{\hat{Y}}}
	[\|\nabla_{\hat{Y}} F (\hat{Y})-k \cdot \mathbf{1}\|_2^2],}
	\end{aligned}
	\label{eq:swgan}
\end{equation}
where $\theta$ is embedded in $D$, and $\mathbf{1}$ is a vector with all entries being 1. We sample $\hat{X}, \hat{Y}$ based on~\cite{gulrajani2017improved}. $\lambda_1, \lambda_2$ are coefficients which balance the penalty terms (see Alg.~\ref{alg:swgan}). For the sake of computational efficiency our objective swaps the order of maximum and integral compared to Eq.~\ref{eq:dualSWD}. This exchange results in a lower bound estimation of Eq.~\ref{eq:dualSWD}. It implies that the objective can lead to the convergence of the dual SWD.

\noindent\textbf{Discussion.}
In addition to the proposed SWAE and SWGAN, it is also possible to apply our proposed primal and dual SWD approximations to other generative models.
For example, following~\cite{makhzani2015adversarial}, AE-based models can be enhanced by adversarial training using our dual SWD.
Inspired by~\cite{salimans2018improving} it is possible to regularize GAN with our primal SWD. Moreover, by using a sorting algorithm~\cite{kolouri2018sliced} we can incorporate a simple primal SWD loss into GAN models. 

\subsection{Training for SWAE \& SWGAN}
To train the proposed SWAE and SWGAN models, we utilize the standard Adam optimization algorithm~\cite{kingma2014adam}.
Throughout training, the projection matrices in the SWD blocks should remain orthogonal.
For this purpose we first initialize the parameters of SWD blocks with random orthogonal matrices through QR decomposition, then update them on a curved manifold instead of a Euclidean space.
Building upon the manifold-valued update rule~\cite{huang2017riemannian}, we optimize the orthogonal matrices on Stiefel manifolds\footnote{A compact Stiefel manifold $St(d,n)$ is defined as $St(d,n) = \{A \in \mathbb{R}^{n \times d}: A^T A = I_d \}$, where $I_d$ is the $d\times d$ identity matrix.}.

In the $t$-th training step, after computing the Euclidean gradient $\nabla L^{(k)}_{\bm{O}_t}$ of an orthogonal matrix $\bm{O}_t$, we obtain its tangential component by subtracting $\bm{O}_t (\bm{O}_t^T \nabla L_{\bm{O}_t}^{(k)} + (\nabla L_{\bm{O}_t}^{(k)})^T \bm{O}_t) / 2$ (see~\cite{boumal2014manopt}), where $L^{(k)}$ is the loss for the $k$-th layer.
For simplicity, we subsequently drop the index $k$. 
Searching along the tangential direction yields the update in the tangent space of the Stiefel manifold.
In the end, the resulting update is projected back to the Stiefel manifold by a retraction operation $\Gamma$. 
Accordingly, the update of the current orthogonal matrix $\bm{O}_t$ on the Stiefel manifold can be written in the following form
\begin{alignat}{2}
	\tilde{\nabla} L_{\bm{O}_t}&=  (\nabla L_{\bm{O}_t}- \bm{O}_t (\nabla L_{\bm{O}_t})^T \bm{O}_t) / 2, \label{eq:otho1} \\
	\bm{O}_{t+1} &= \Gamma(\bm{O}_t - \Omega(\tilde{\nabla} L_{\bm{O}_t})), 
	\label{eq:otho2}
\end{alignat}
where $\Gamma$ denotes the retraction operation corresponding to QR decomposition and $\Omega(\cdot)$ denotes the standard Adam optimization. 
Note that the retraction has complexity $O(r^3)$ for $r$-dimensional data and is the main contributor to the time complexity of the optimization method. 
We therefore encode the $n$-dimensional input data into $r$-dimensional latent codes ($r < n$) before applying the SWD blocks, such that the training speed of our method remains comparable to the existing methods (see Tab.~\ref{tab:vae_gan_results}).
The inference speed is not affected by the retraction operation.

\section{Experiments under Standard Training}
After discussing the theoretical merits of SWD and its applications to generative modeling, we examine the practical advantages of our proposed models under a standard training setting.

\begin{figure}[h!]
\resizebox{\linewidth}{!}{%
\centering
\scriptsize
\begin{tabular}{@{\hskip -0.07in}c@{\hskip -0.07in}c@{\hskip -0.07in}c@{\hskip -0.07in}c@{\hskip -0.07in}c@{\hskip -0.07in}c@{\hskip -0.07in}}
\toprule

WAE & AE +100 IDT & \textbf{SWAE} & CT-GAN &SWG & \textbf{SWGAN}\\ \midrule
\includegraphics[width=0.18\linewidth]{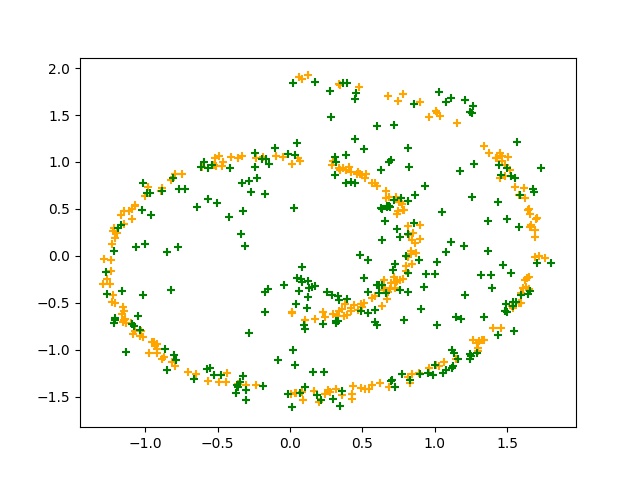}&
\includegraphics[width=0.18\linewidth]{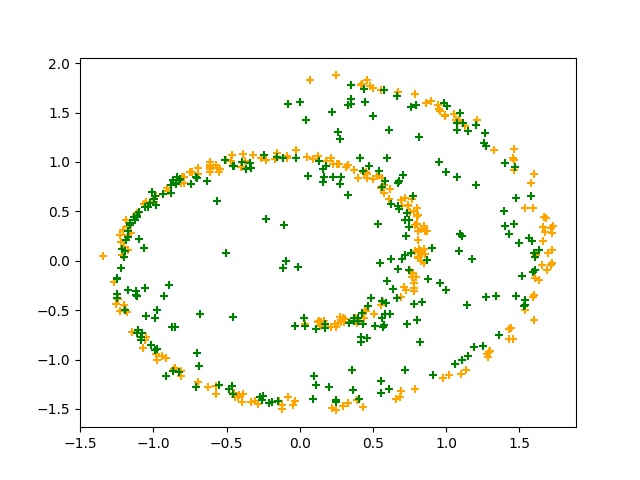}&
\includegraphics[width=0.18\linewidth]{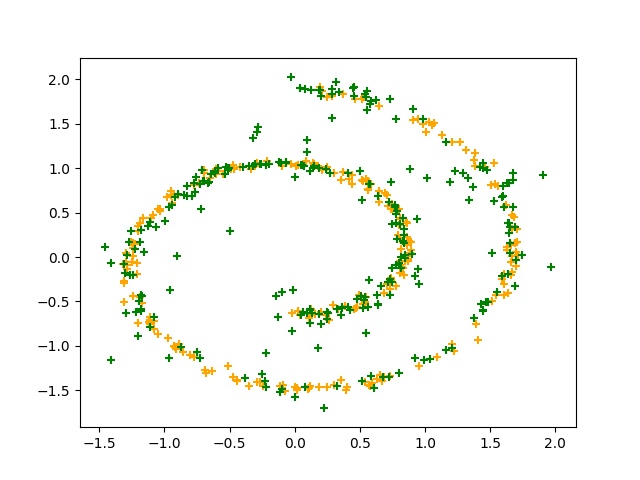}&
\includegraphics[width=0.18\linewidth]{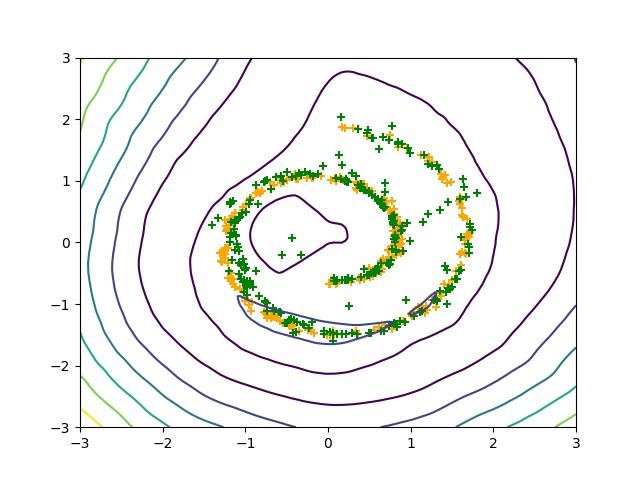}&
\includegraphics[width=0.18\linewidth]{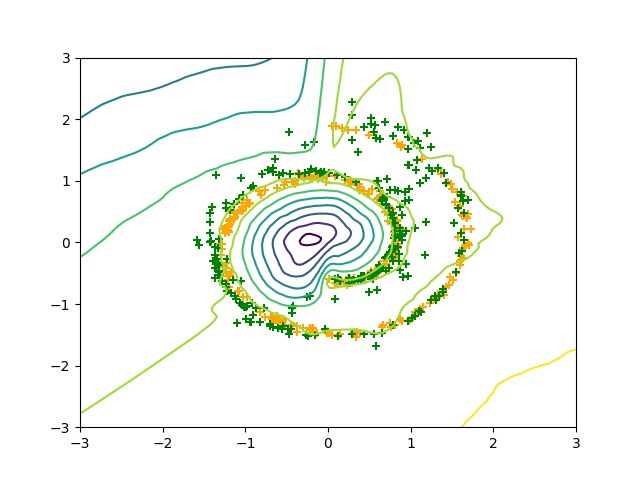}&
\includegraphics[width=0.18\linewidth]{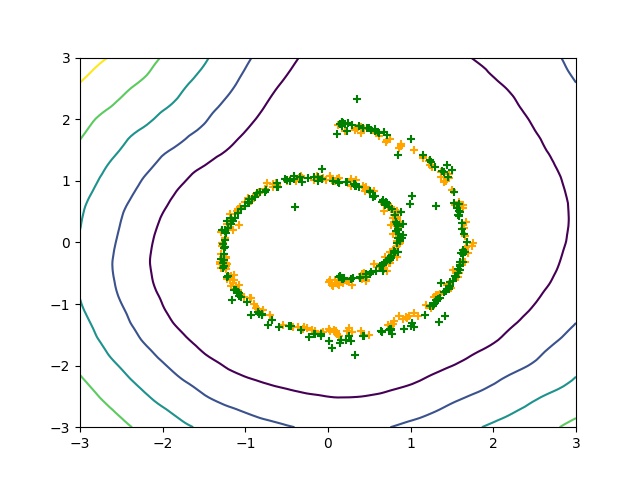}\\

0.04 (0.06) & 0.05 (0.06) & \textbf{0.04 (0.04)} & 0.01 & 0.03 & \textbf{0.01} \\

\includegraphics[width=0.18\linewidth]{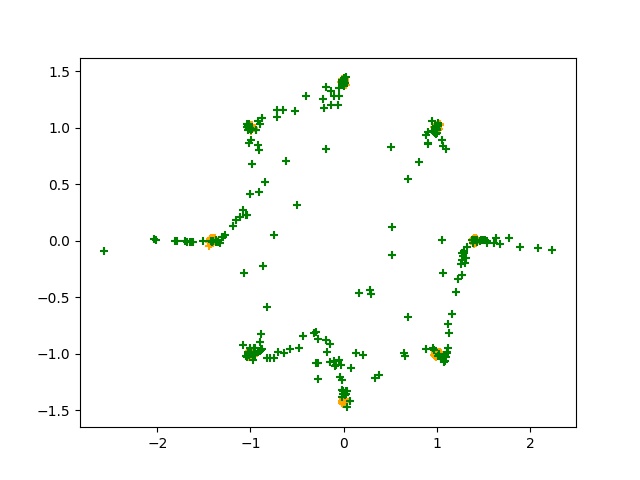}&
\includegraphics[width=0.18\linewidth]{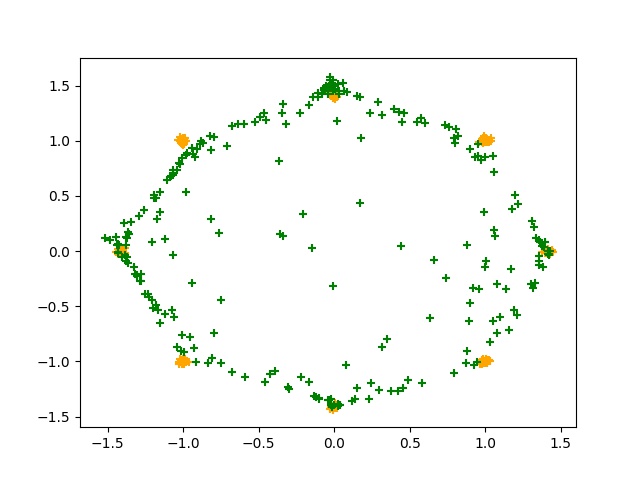}&
\includegraphics[width=0.18\linewidth]{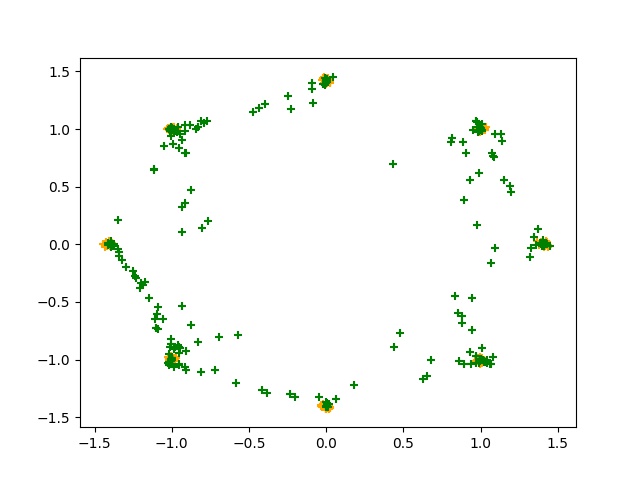}&
\includegraphics[width=0.18\linewidth]{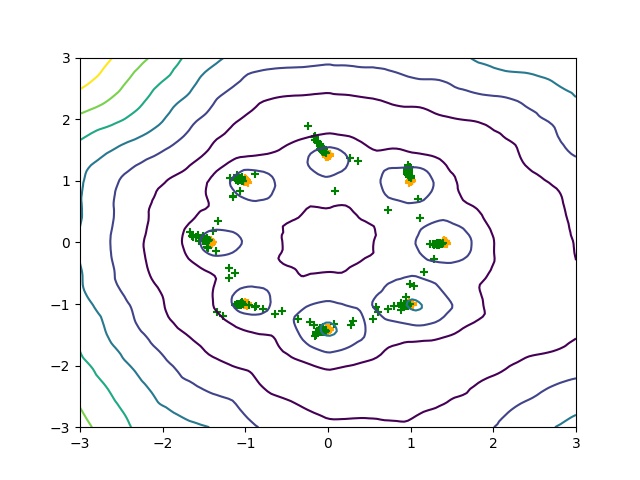}&
\includegraphics[width=0.18\linewidth]{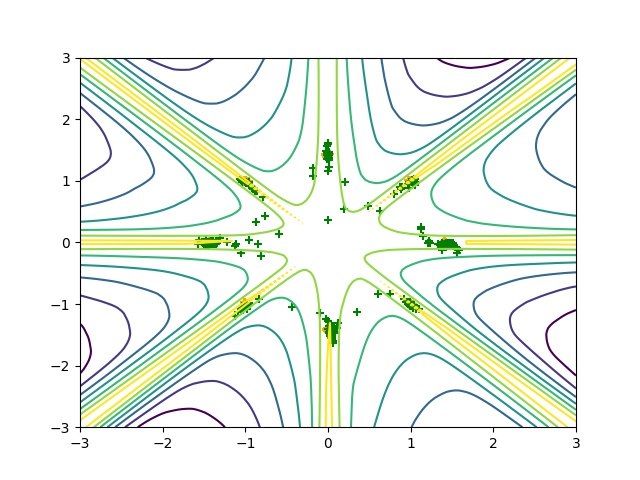}&
\includegraphics[width=0.18\linewidth]{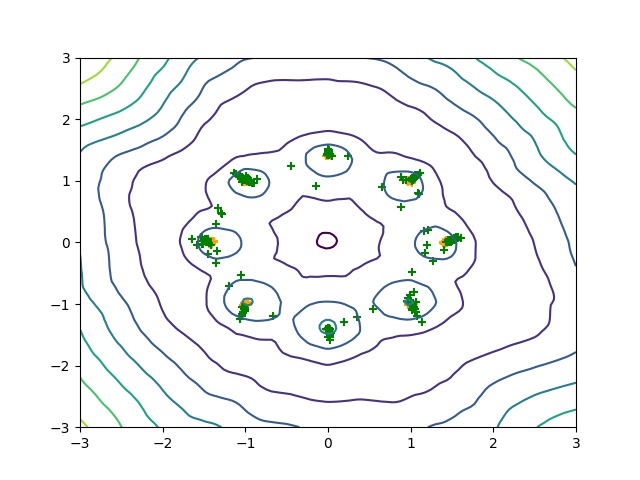}\\

0.07 (0.04) & 0.06 (0.06) & \textbf{0.03 (0.02)} & 0.03 & 0.05 & \textbf{0.02} \\

\includegraphics[width=0.18\linewidth]{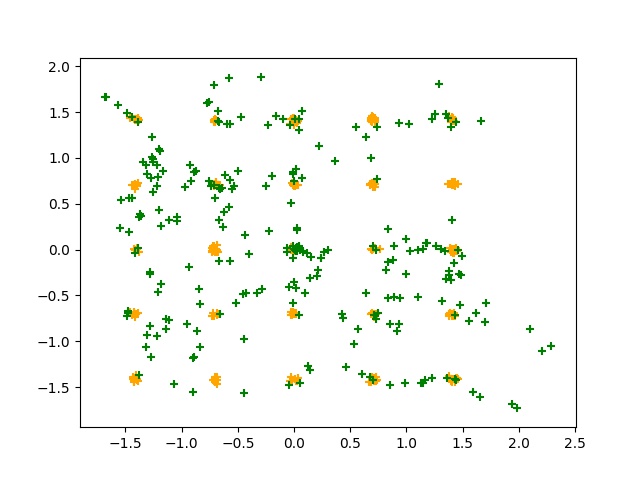}&
\includegraphics[width=0.18\linewidth]{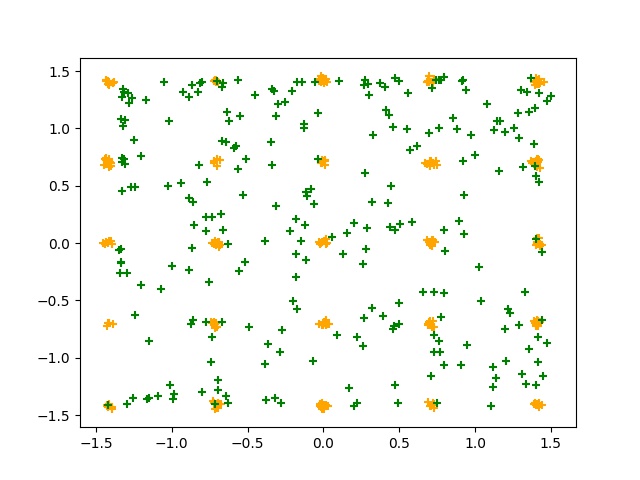}&
\includegraphics[width=0.18\linewidth]{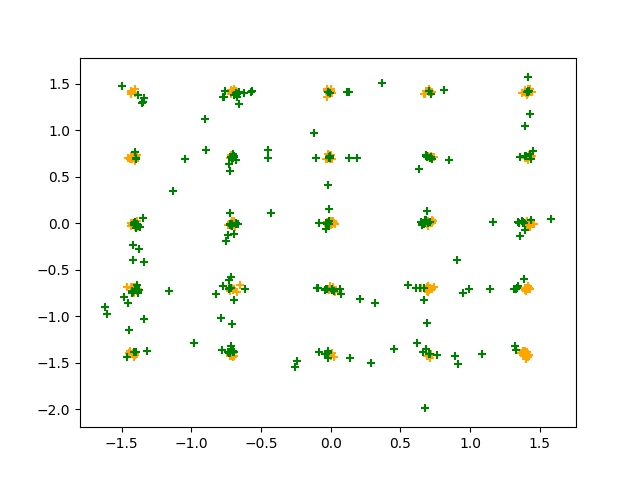}&
\includegraphics[width=0.18\linewidth]{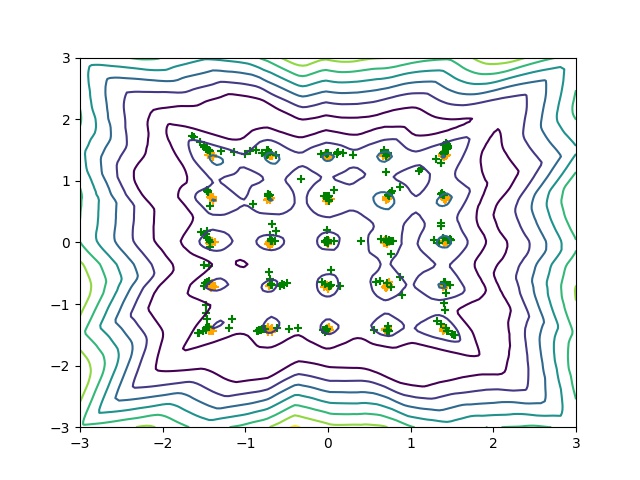}&
\includegraphics[width=0.18\linewidth]{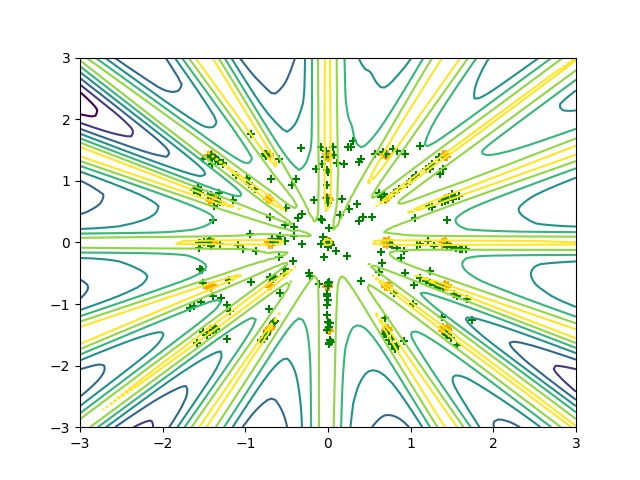}&
\includegraphics[width=0.18\linewidth]{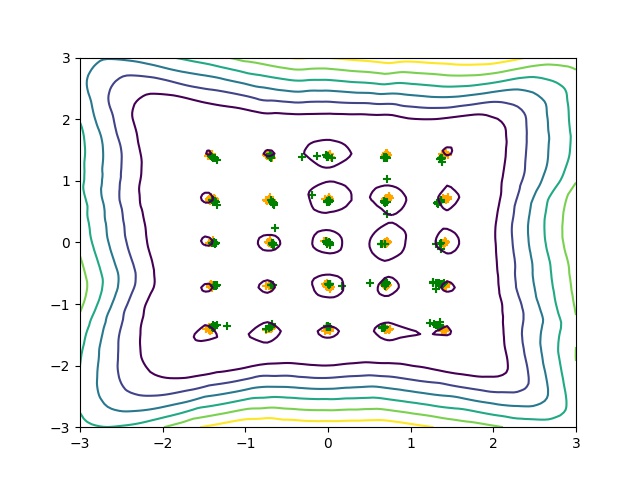}\\ 

0.05 (0.03) & 0.05 (0.06) & \textbf{0.04 (0.02)} & 0.02 & 0.05 & \textbf{0.01} \\ \bottomrule

\end{tabular}
}
\caption{Visual and FID results for generated samples (green dots) compared to real samples (yellow dots) on Swiss Roll (top row), 8 Gaussians (middle row) and 25 Gaussians (bottom row).
For the AE-based models, the FID scores displayed in parentheses indicate the discrepancy between generated latent codes and real noise. For the GANs, value surfaces of the discriminators are also plotted.}
\label{fig:toy}
\end{figure}

\subsection{Evaluation on Toy Datasets}
Following~\cite{gulrajani2017improved}, 
we first conduct experiments on three toy datasets: Swiss Roll, 8 Gaussians and 25 Gaussians (see Fig.~\ref{fig:toy}).
For a fair comparison, we respect the experimental settings of~\cite{gulrajani2017improved} for the compared methods.
For this experiment, the SWAE and SWGAN use only one SWD block.
The superiority of our models is illustrated by visual results and the Fr\'echet inception distance (FID)~\cite{heusel2017gans}.

\noindent \textbf{SWD vs WD.}
Compared to the WD-based models--WAE and CTGAN--Fig.~\ref{fig:toy} shows that our SWAE outperforms WAE both visually and quantitatively. SWGAN also achieves better scores than CTGAN. 
These results support the advantages of SWD over WD for generative modeling.

\noindent \textbf{SWAE vs AE + IDT.}
IDT~\cite{pitie2007automated} is the starting point of our SWAE.
Thus, we also use it as a baseline. 
For this purpose,  we stack IDT blocks on top of a regular encoder. We determine the optimal number of IDT blocks to be $100$ by running multiple experiments. Then we train the IDT enhanced AE (AE + 100 IDT) under the standard Adam optimization. 
The FID scores in parentheses in Fig.~\ref{fig:toy} indicate that our SWAE, equipped with only one SWD block,
better approximates the real noise distribution than the IDT enhanced AE with 100 IDT blocks. 
Moreover, the visual results confirm the improvement of SWAE over AE + 100 IDT for all three datasets.
This shows that a single learnable primal SWD block is more effective than multiple original IDT blocks.
 
\noindent \textbf{SWGAN vs SWG.}
We compare our SWGAN to the state-of-the-art SWD-based GAN model SWG~\cite{deshpande2018generative}.
SWG is a typical application of SWD approximation with projections along random unit vectors. 
Despite the use of 10000 random unit vectors in SWD, Fig.~\ref{fig:toy} shows that our SWGAN with only one dual SWD block (128 orthogonal unit vectors) is more successful at capturing the real data distribution in terms of better visual and better FID results, 
leveraging learnable projections along orthogonal unit vectors.
\begin{table}
\resizebox{\linewidth}{!}{%
\centering
\scriptsize
\begin{tabular}{c|ccc|cc}
\toprule
             & CIFAR-10        & CelebA         & LSUN   & CIFAR-10   &  CelebA        \\ \midrule
VAE           & 144.7$\pm$9.6          & 66.8$\pm$2.2        &  --  & 0.16s & 0.64s \\
WAE-MMD       & 109.1$\pm$1.5          & 59.1$\pm$4.9    &   --     & 0.17s & 0.63s\\
AAE (WAE-GAN)  & \textbf{107.7$\pm$2.1}          & \textbf{49.3$\pm$5.8}  &    --   & 0.25s & 1.61s  \\
\textbf{SWAE} & \textbf{107.9$\pm$5.2} & \textbf{48.9$\pm$4.3} & --  & 0.16s & 0.37s \\ \midrule
DCGAN          & 30.2 $\pm$ 0.9           & 52.5 $\pm$ 2.2            & 61.7 $\pm$ 2.9    & 0.13s     &1.57s \\
WGAN           & 51.3 $\pm$ 1.5           & 37.1 $\pm$ 1.9           & 73.3 $\pm$ 2.5  & 0.25s       &2.12s  \\
WGAN-GP        & 19.0  $\pm$ 0.8        & 18.0 $\pm$ 0.7         & 26.9 $\pm$ 1.1    & 0.60s & 2.40s        \\
SNGAN        & 21.5 $\pm$ 1.3         & 21.7  $\pm$ 1.5       & 31.3 $\pm$ 2.1 &0.21s   & 0.53s \\
CTGAN        & 17.6  $\pm$ 0.7        & 15.8  $\pm$ 0.6        & 19.5 $\pm$ 1.2  & 0.63s & 2.61s \\
SWG            & 33.7  $\pm$ 1.5        & 21.9 $\pm$ 2.0         & 67.9 $\pm$ 2.7     & 0.22s & 0.83s       \\
\textbf{SWGAN} & \textbf{17.0 $\pm$ 1.0} & \textbf{13.2 $\pm$ 0.7} & \textbf{14.9 $\pm$ 1.0} & 0.64s & 2.74s \\ \bottomrule
\end{tabular}
}
\caption{FID (left) and runtime (right) comparison of AE-based and GAN models. The runtime is computed for one training step on a TITAN Xp GPU. }
\label{tab:vae_gan_results}
\vspace{-0.5cm}
\end{table}

\subsection{Evaluation on Standard Datasets}
In addition to our toy dataset experiments, we also conduct various experiments on three widely-used benchmarks: CIFAR-10~\cite{krizhevsky2009learning}, CelebA~\cite{liu2015deep}, and LSUN~\cite{yu2015lsun}. 
We compare our SWAE to VAE~\cite{kingma2013auto}, WAE-MMD~\cite{tolstikhin2017wasserstein}, and AAE (WAE-GAN)~\cite{tolstikhin2017wasserstein,makhzani2015adversarial}. For GAN models,
we compare our SWGAN against DCGAN~\cite{radford2015unsupervised}, WGAN~\cite{arjovsky2017wasserstein}, WGAN-GP~\cite{gulrajani2017improved},
SNGAN~\cite{miyato2018spectral},
CTGAN~\cite{wei2018improving}, and SWG~\cite{deshpande2018generative}.

Our proposed SWAE uses the decoder architecture suggested by~\cite{berthelot2017began}.
For the encoder, we stack our primal SWD blocks on a shallow encoding network containing a downscaling and linear transform layer.
Our SWGAN employs the ResNet structure used by ~\cite{gulrajani2017improved} for the generator.
For the discriminator, we apply our dual SWD blocks to an encoding network containing multiple ResNet layers.
Please refer to our supplementary material for more architecture details.
As to the compared methods, we use the official implementation if it is available online, and we apply the optimal settings tuned by their authors. 

\begin{figure*}[t!]
\scriptsize
\includegraphics[width=0.65\linewidth, height=3.6cm]{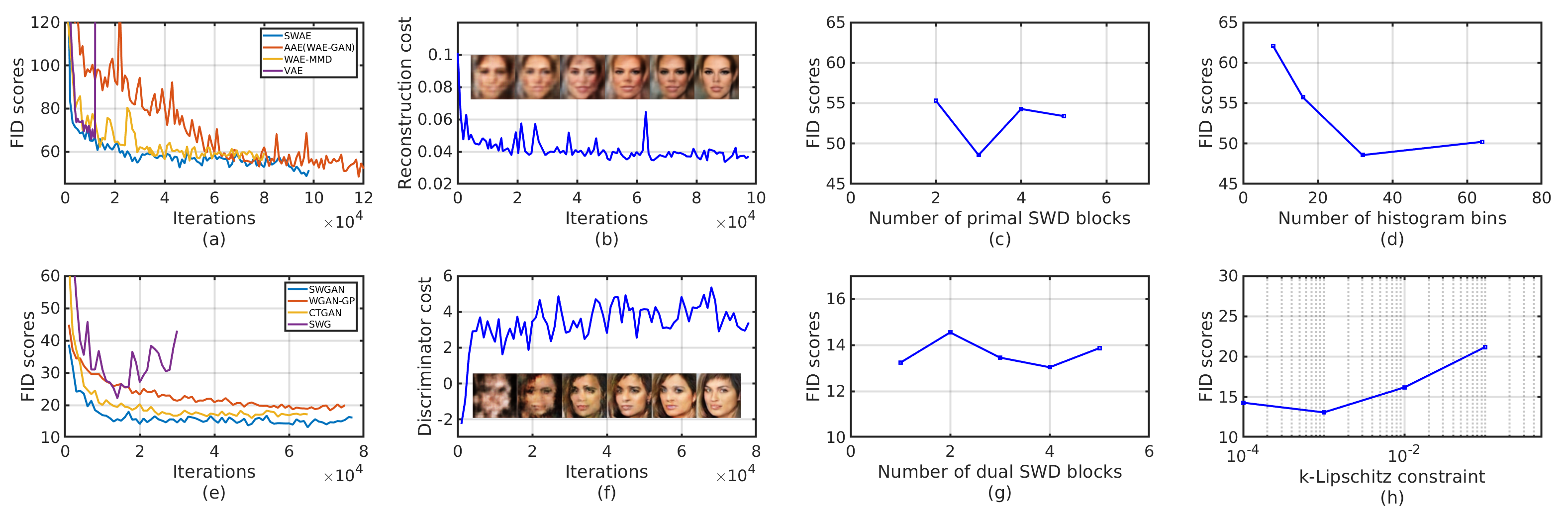}
\begin{tabular}[b]{ll}
\hline
\multicolumn{2}{c}{Human Preference Score} \\ \hline
\textbf{SWAE} / VAE                 & \textbf{0.74} / 0.26   \\
\textbf{SWAE} / WAE-MMD             & \textbf{0.55} / 0.45   \\
\textbf{SWAE} / AAE (WAE-GAN)        &  0.48 / \textbf{0.52}   \\ \hline
\textbf{SWGAN} / WGAN-GP            & \textbf{0.61} / 0.39   \\
\textbf{SWGAN} / SNGAN              & \textbf{0.66} / 0.34   \\
\textbf{SWGAN} / CTGAN              & \textbf{0.56} / 0.44   \\
\textbf{SWGAN} / SWG                & \textbf{0.69} / 0.31   \\ \hline
\textbf{PG-SWGAN} / PG-WGAN         & \textbf{0.55} / 0.45   \\ \hline
\textbf{PG-SWGAN-3D} / VGAN         & \textbf{0.91} / 0.09   \\
\textbf{PG-SWGAN-3D} / MoCoGAN      & \textbf{0.95} / 0.05   \\
\textbf{PG-SWGAN-3D} / PG-WGAN-3D   & \textbf{0.54} / 0.46   \\
\hline
\end{tabular}

\caption{Left: Training and hyperparameter study for SWAE (top row) and SWGAN (bottom row). Right: Preference score from MTurk user study for generated images on CelebA and synthesized videos on TrailerFaces. }
\label{fig:swg_curve}
\end{figure*}

\begin{figure*}[t!]
\scriptsize
\centering
\resizebox{1\linewidth}{!}{%
\begin{tabular}{ccccc}
\toprule
& VAE & WAE-MMD & AAE (WAE-GAN) & \textbf{SWAE}\\ \hline
CIFAR-10 & \includegraphics[width=0.15\linewidth]{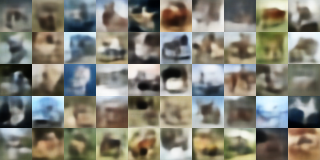}&
\includegraphics[width=0.15\linewidth]{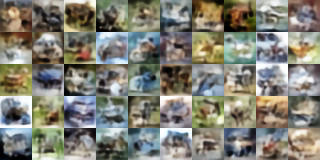}&
\includegraphics[width=0.15\linewidth]{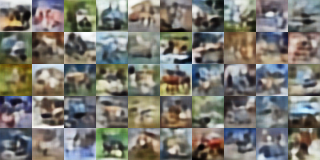}&
\includegraphics[width=0.15\linewidth]{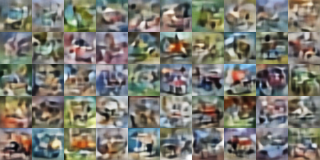}\\

CelebA & \includegraphics[width=0.15\linewidth]{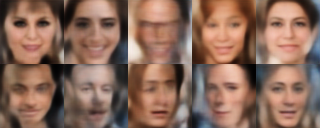}&
\includegraphics[width=0.15\linewidth]{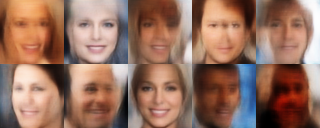}&
\includegraphics[width=0.15\linewidth]{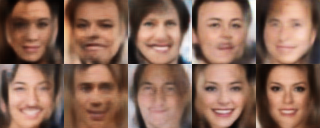}&
\includegraphics[width=0.15\linewidth]{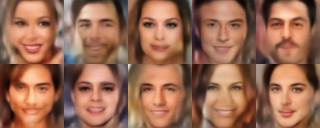}\\\bottomrule

 & SWG & CTGAN & WGAN-GP & \textbf{SWGAN}\\ \hline
CIFAR-10 & \includegraphics[width=0.15\linewidth]{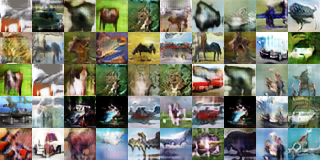}&
\includegraphics[width=0.15\linewidth]{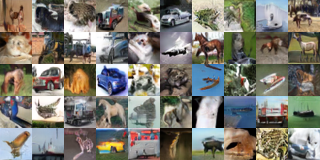}&
\includegraphics[width=0.15\linewidth]{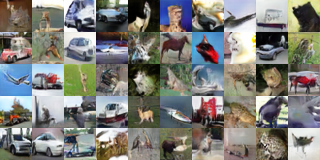}&
\includegraphics[width=0.15\linewidth]{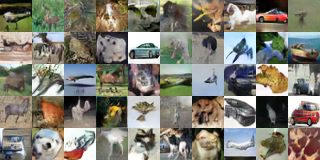}\\ 

CelebA & \includegraphics[width=0.15\linewidth]{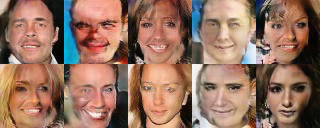}&
\includegraphics[width=0.15\linewidth]{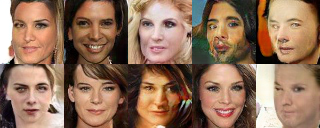}&
\includegraphics[width=0.15\linewidth]{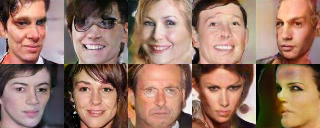}&
\includegraphics[width=0.15\linewidth]{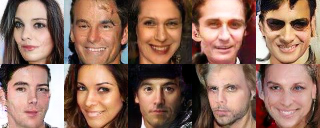}\\

LSUN & \includegraphics[width=0.15\linewidth]{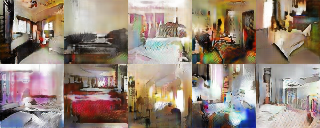}&
\includegraphics[width=0.15\linewidth]{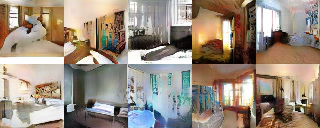}&
\includegraphics[width=0.15\linewidth]{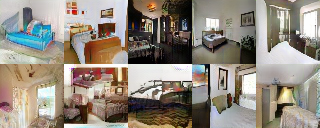}&
\includegraphics[width=0.15\linewidth]{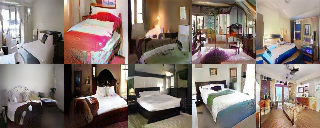}\\
\bottomrule
\end{tabular}
}
     \caption{Visual results for SWAE, SWGAN, and compared methods. More results are available in the supplementary material.}
	\label{fig:swae_swgan_visual}
\vspace{-0.195cm}
\end{figure*}

The evaluations of AE-based models in Tab.~\ref{tab:vae_gan_results}, Fig.~\ref{fig:swg_curve} (Right: MTurk preference score) and Fig.~\ref{fig:swae_swgan_visual} show that our proposed SWAE clearly outperforms the pure VAE model.
Furthermore, our FID score is only marginally higher than the one of AAE (WAE-GAN), which additionally employs adversarial training.
Due to this adversarial training, AAE (WAE-GAN) is generally less stable,
while our model provides stable training (see Fig.~\ref{fig:swg_curve} (Left a)) owing to a simple $l_2$ reconstruction loss without any additional regularizer. 

Tab.~\ref{tab:vae_gan_results} and Fig.~\ref{fig:swg_curve} (Right) highlight the advantages of our SWGAN model in terms of FID and MTurk preference score. 
Relying on extra label information, SNGAN achieved a competitive FID score of 17.5 on CIFAR-10 as reported in~\cite{miyato2018spectral}.
Meanwhile, our SWGAN reaches an even lower score of 17.4 without the use of ground truth labels.
The visual results reported in Fig.~\ref{fig:swae_swgan_visual} are consistent with the FID scores in Tab.~\ref{tab:vae_gan_results}. 
We believe that the good performance of SWGAN mainly results from the efficient approximation of the proposed SWD on multiple one-dimensional marginal distributions of the training data.

\begin{table}[h!]
\centering
\resizebox{\linewidth}{!}{
\begin{tabular}{ccccc}
\toprule
& ResNet (w/ norm) & ResNet (w/o norm) & ConvNet (w/ norm) & ConvNet (w/o norm) \\ \hline
WAE-MMD &64.0   &61.8 &55.8 &67.8 \\

AAE (WAE-GAN)& \textbf{62.3}  &\textbf{56.7}	& \textbf{48.3} &66.1	 \\ 
 
\textbf{SWAE} & 63.2   & 59.1  &65.2 & \textbf{48.6}           \\ 
 \bottomrule
CTGAN  & 16.0       &16.5	&19.5	&19.7            \\ 
SWG & 24.3 & 29.1 & 22.2 &28.5 \\ 
\textbf{SWGAN} & \textbf{13.0}  & \textbf{14.8}  & \textbf{19.2} & \textbf{18.8}  \\ 

\bottomrule

\end{tabular}
}

\caption{FID scores of various architectures on CelebA. 
The optimal architectures are ConvNet for SWG, WAE-MMD, AAE (WAE-GAN), ResNet for CTGAN, SWGAN, and ConvNet without normalization (w/o norm) for SWAE.}
\label{tab:stable}
\vspace{-0.18cm}
\end{table}

\noindent\textbf{Stability Study. }
In addition to the optimal architecture comparison, we also study the model stability under a variety of settings: ConvNet and ResNet, with normalization (w/ norm) and without normalization (w/o norm). As shown in Tab.~\ref{tab:stable}, our proposed models are less sensitive in terms of FID scores, which is credited to the easier approximation of SWD (see visual results in supplementary material).

\noindent\textbf{Training and Hyperparameters.} Fig.~\ref{fig:swg_curve} (Left b, f) show that the visual quality produced by our SWAE and SWGAN increases progressively with increasing training iterations. 
Also, using a small number of SWD blocks---$3$ primal and $4$ dual blocks (Fig.~\ref{fig:swg_curve} (Left c, g))---is sufficient to achieve top performances.
The $4$ dual blocks ($4 \times 128$ unit vectors) in our SWGAN, stand in contrast to the 10000 random unit vectors required by SWG~\cite{deshpande2018generative}.
This confirms the efficiency of our learnable SWD blocks. 
In another experiment, we study the impact of the Lipschitz constant $k,k'$ for SWGAN and bin number $l$ for SWAE.
Fig.~\ref{fig:swg_curve} (Left d, h) show that SWGAN favors relatively small values of $k$. $k'$ is determined to be 0 (see supplementary material). The optimal bin number for SWAE is $32$. 
By analyzing the trade-off between computational complexity and performance, we set the $r$-dimension to $128$ (Alg.~\ref{alg:SOT},~\ref{alg:dualSOT}).
We also determine $\lambda_1, \lambda_2$ (Eq.~\ref{eq:swgan}) to be $20, 10$ using grid search. 
Both studies are presented in the supplementary material.
The learning rate of SWGAN and SWAE is determined empirically to be $0.0003$.
Finally, we set the discriminator iterations per training step of SWGAN to $4$ for LSUN and CelebA and $5$ for CIFAR-10.

\section{Experiments under Progressive  Training}

Encouraged by the visual quality and stability improvements on standard benchmarks, we evaluate our proposed model for high resolution image and video generation under the progressive training manner suggested by \cite{karras2017progressive}.

\noindent\textbf{Higher Resolution Image Generation.} 
For this task, we use the CelebA-HQ~\cite{karras2017progressive}  and LSUN~\cite{yu2015lsun} datasets, which contain 1024$\times$1024 and 256$\times$256 images respectively. To improve high resolution image generation, \cite{karras2017progressive} introduces a progressive growing training scheme for GANs (PG-GAN). PG-GAN uses WGAN-GP loss (PG-WGAN) to achieve state-of-the-art high resolution image synthesis. For fair comparison, we equip the same progressive growing architecture with our proposed SWGAN objective and its dual SWD blocks (PG-SWGAN). As shown in Fig.~\ref{fig:swg_curve} (Right) and Fig.~\ref{fig:pgtraining_visual}, our PG-SWGAN can outperform PG-WGAN in terms of both qualitative and quantitative comparison on the CelebA-HQ and LSUN datasets.

\noindent\textbf{Higher Resolution Video Generation.} We introduce a new baseline unsupervised video synthesis method along with a new facial expression video dataset\footnote{Both the baseline code and dataset will be released at \url{https://github.com/musikisomorphie/swd.git}}. The dataset contains approximately 200,000 individual clips of various facial expressions, where the faces are cropped with $256\times256$ resolution from about 6,000 Hollywood movie trailers on YouTube. Thus, we name the dataset as TrailerFaces. For progressive video generation, we exploit a new PG-GAN network design for unsupervised video generation. We progressively scale the network in spatio-temporal dimension such that it can produce spatial appearance and temporal movement smoothly from coarse to fine.  Fig.~\ref{fig:swg_curve} (Right) shows the superiority of our model over the state-of-the-art methods \cite{vondrick2016generating,tulyakov2017mocogan} in terms of preference score. For qualitative comparison, please refer to our supplementary videos.

Based on the proposed PG-GAN design, we evaluate the original WGAN loss (PG-WGAN-3D) and our proposed SWGAN loss (PG-SWGAN-3D). Fig.~\ref{fig:swg_curve} (Right) and Fig.~\ref{fig:pgtraining_visual} present their qualitative and quantitative comparison. For the FID evaluation, we follow~\cite{wang2018video} to compute the video FID scores for PG-WGAN-3D and PG-SWGAN-3D. The higher preference score (Fig.~\ref{fig:swg_curve} (Right)) and the lower FID score (Fig.~\ref{fig:pgtraining_visual}) of our PG-SWGAN-3D reflect the advantage of using our proposed SWGAN.

\noindent\textbf{Limitations.}
For pure AE-based generative models, including SWAE, the extension to high resolution image and video synthesis tasks is non-trivial. The challenges for SWAE to generate high quality images and videos on par with SWGAN remain. We plan to address this performance gap in future research.

\begin{figure*}[t!]
\centering
\scriptsize
\resizebox{1\linewidth}{!}{%
\begin{tabular}{ccc}
\toprule

& PG-WGAN & \textbf{PG-SWGAN}\\ \hline

CelebA-HQ & \includegraphics[width=0.4\linewidth]{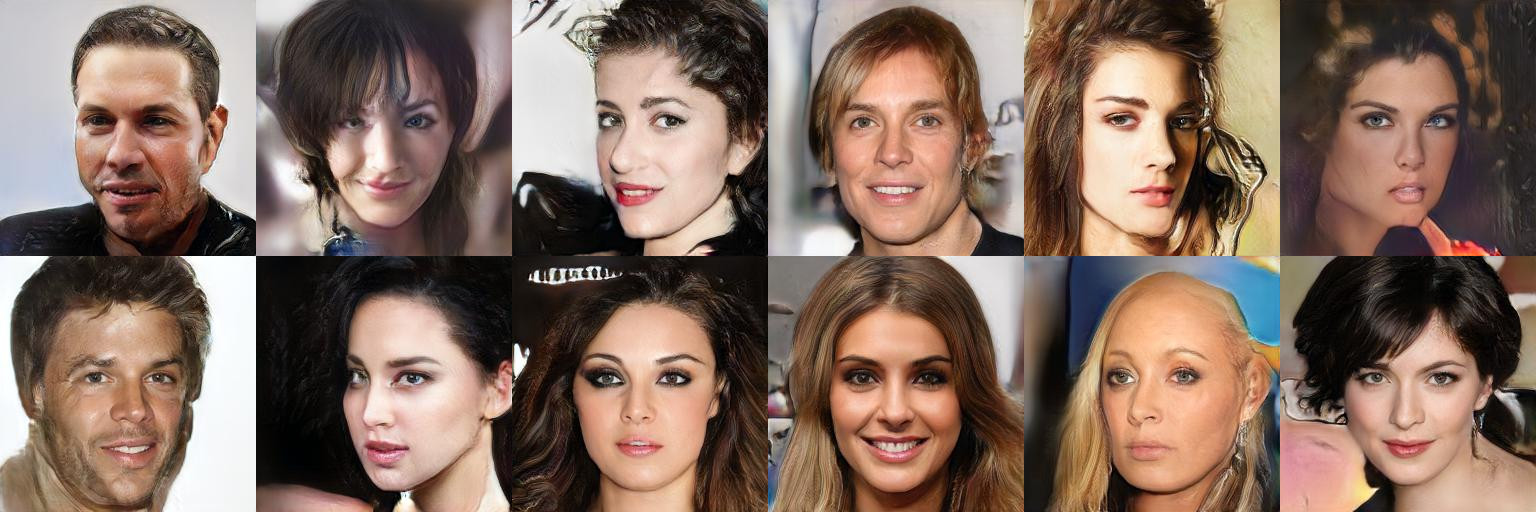} &
\includegraphics[width=0.4\linewidth]{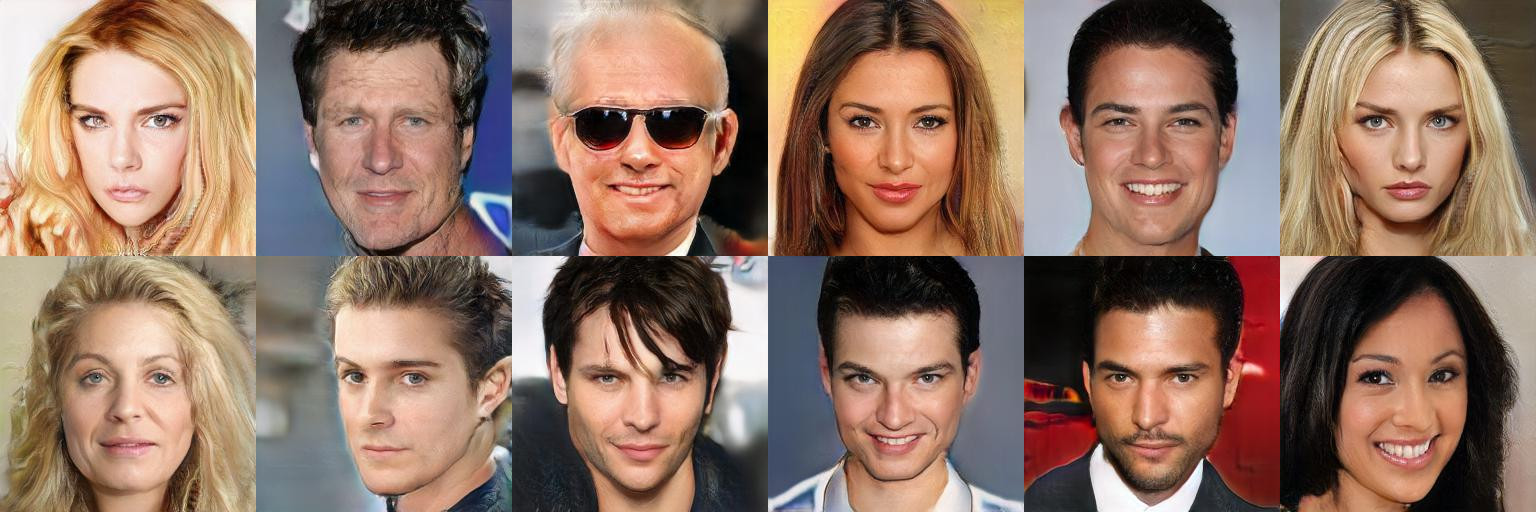}\\

& 7.5 & \textbf{5.5} \\

LSUN & \includegraphics[width=0.4\linewidth]{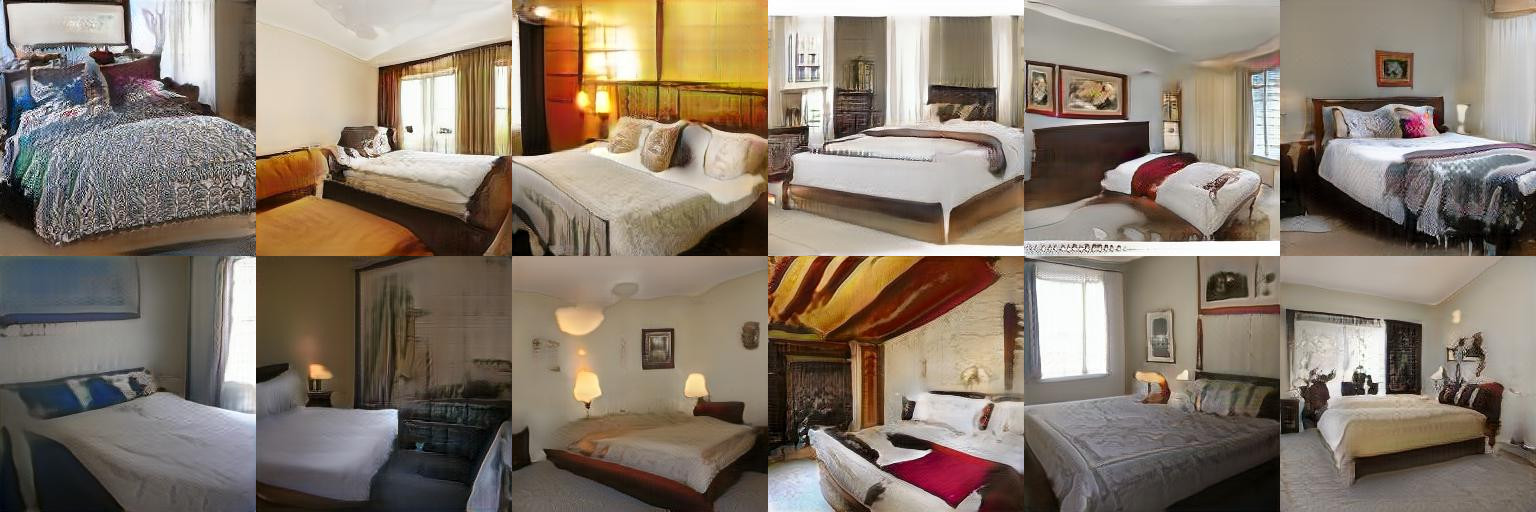}&
\includegraphics[width=0.4\linewidth]{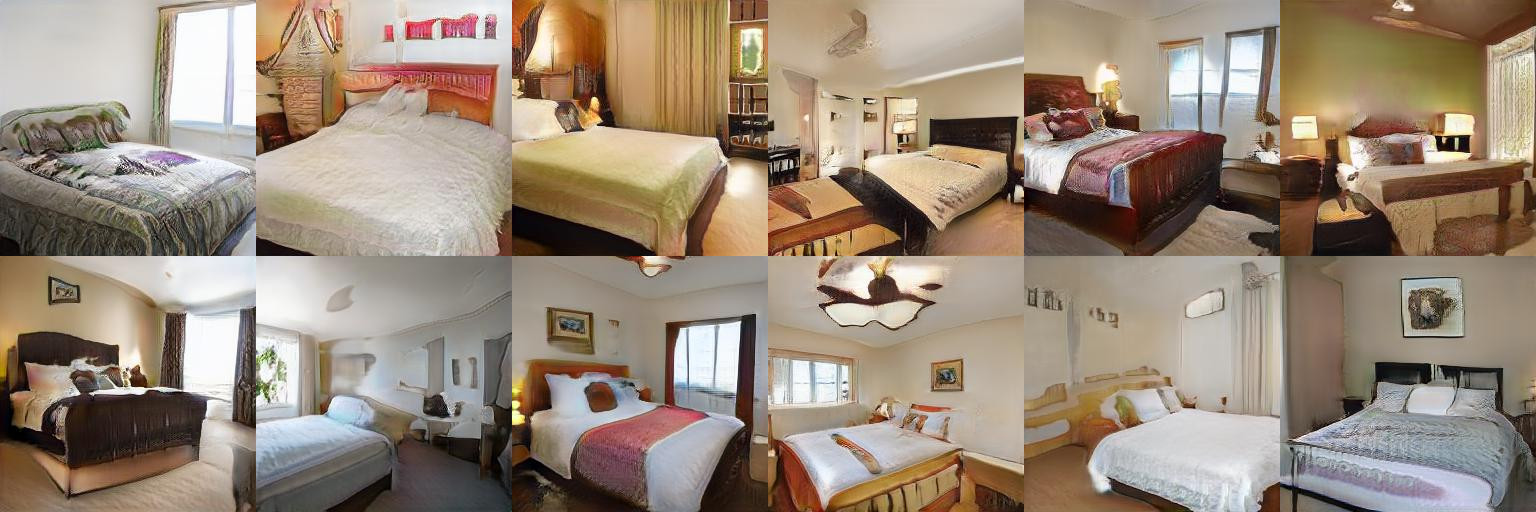}\\
& 8.4 & \textbf{8.0} \\
\toprule
& PG-WGAN-3D & \textbf{PG-SWGAN-3D}\\ \hline
TrailerFaces  & \includegraphics[width=0.4\linewidth]{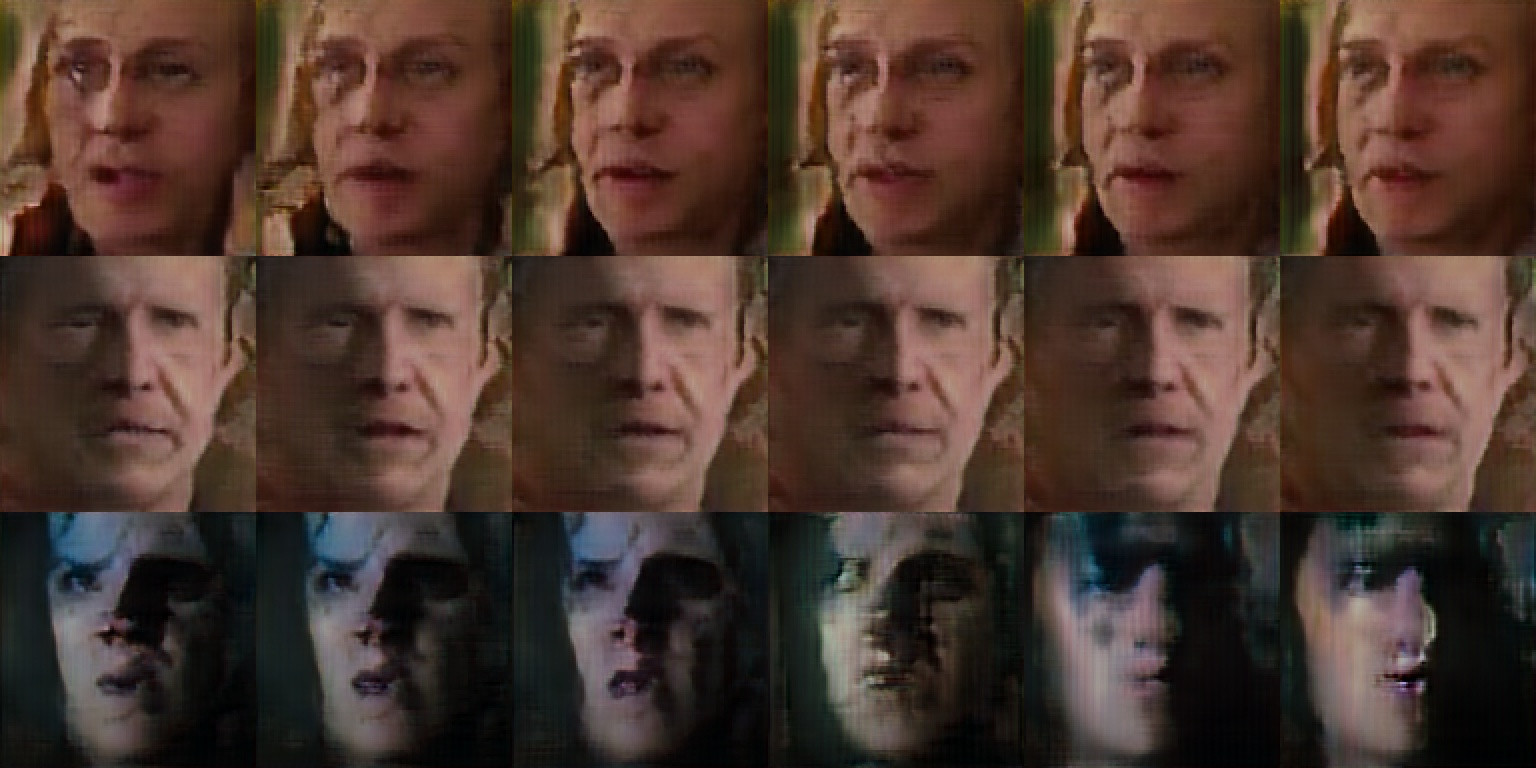} &
\includegraphics[width=0.4\linewidth]{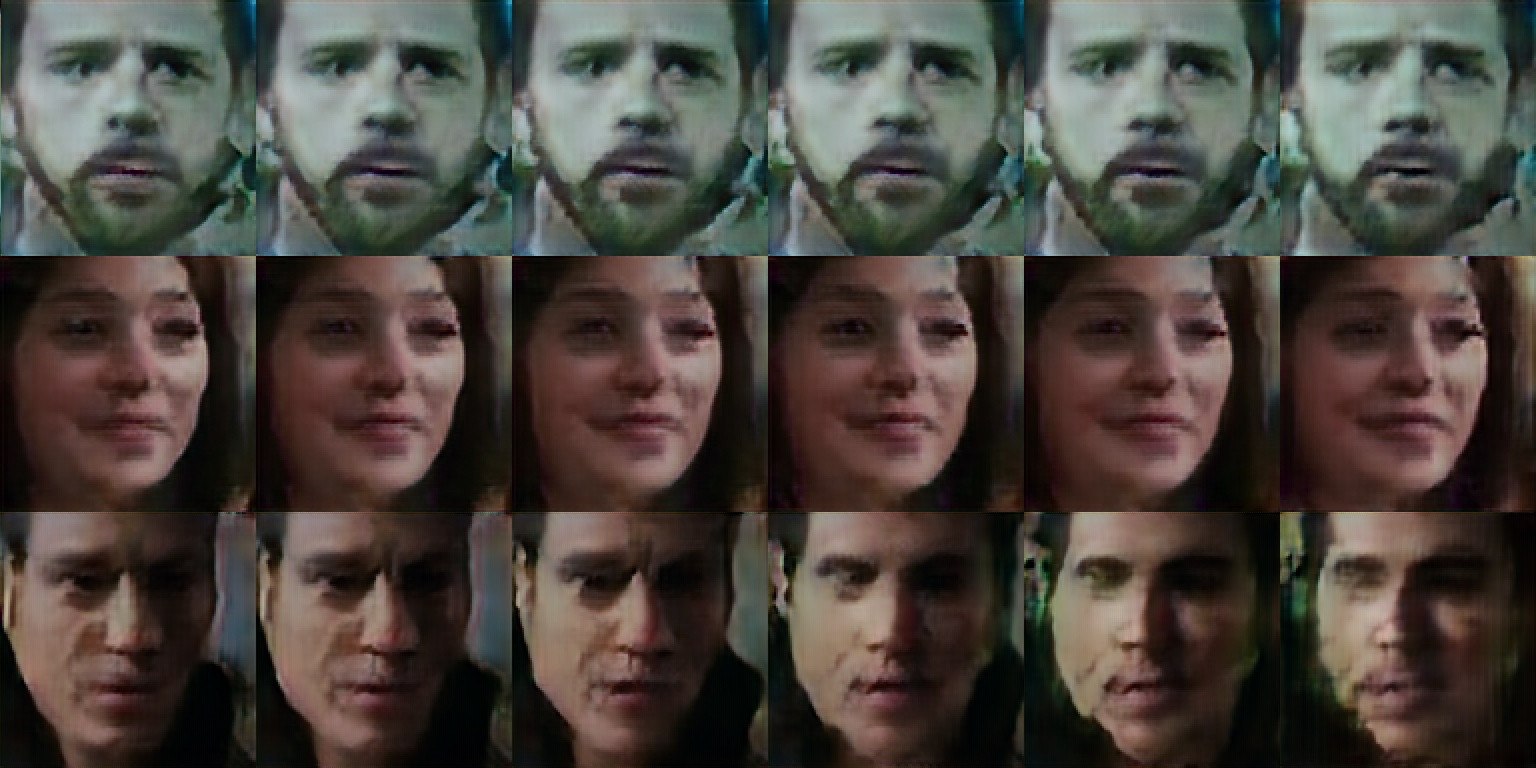}\\

& 462.6 & \textbf{404.1} \\

\bottomrule
\end{tabular}
}
\caption{Visual and FID results for compared methods on higher resolution images/videos. More results are available in the supplementary.}
\label{fig:pgtraining_visual}
\vspace{-0.5cm}
\end{figure*}

\section{Conclusion}
In this paper, we introduce a novel way of efficiently approximating the primal and dual SWD. As concrete applications, we enhance modern AE-based and GAN models with the resulting primal and dual SWD blocks. 
For image and video synthesis, both qualitative and quantitative results show the superiority of our models over other approaches.


\newpage
\onecolumn

\section{Supplementary Material}
\beginsupplement

\subsection*{A \quad Proof of Theorem 1}
In order to prove Theorem 1, we need to apply the Dvoretzky–Kiefer–Wolfowitz inequality,

\begin{thm1}(DKW-inequality)
	Given $b \in \mathbb{N}$, let $Z_1, Z_2, \dots , Z_b $ be real-valued i.i.d.\ random variables with a continuous CDF $F_Z$.  Then we define the associated EDF $F_{Z,b}(t) = \frac{1}{b} \sum_{i=1}^b \mathbf{1}_{\{Z_i \leq t\}}$,
	then for all $\varepsilon \geq\sqrt{\tfrac{1}{2b}\ln2}$ it holds 
	\begin{equation}
		\Pr\Bigl(\| F_{Z,b}(t) - F_{Z}(t) \|_{\infty} > \varepsilon \Bigr) \le e^{-2b\varepsilon^2}. 
	\end{equation} 
\end{thm1}

Based on the DKW-inequality we have the error estimation for Alg. 1:
\begin{thm2}
	\label{thm:theorem1}
	Given $b \in \mathbb{N}$, let $Z_1, Z_2, \dots , Z_b $ be real-valued i.i.d.\ random variables with a continuous CDF $F_Z^{-1}$ with domain $[0,1]$.  Then we define the associated EDF $F^{-1}_{Z,b}(t) = \frac{1}{b} \sum_{i=1}^b \mathbf{1}_{\{Z_i \leq t\}}$.
	Assume $\tilde{F}_Y, F_Y$ are CDFs satisfying $\|\tilde{F}_Y - F_Y \|_{\infty} \leq \gamma$,
	then there exists a $\delta > 0 $ such that for all $\varepsilon- \delta \gamma \geq\sqrt{\tfrac{1}{2b}\ln2}$ it holds 
	\begin{equation}
		\Pr\Bigl(\| F^{-1}_{Z,b}   \tilde{F}_Y (t) - F^{-1}_{Z}   F_Y (t)\|_{\infty} > \varepsilon \Bigr) \le e^{-2b(\varepsilon- \delta \gamma)^2}. 
	\end{equation} 
\end{thm2}
\paragraph{\textit{Proof}}
Let $\tilde{Z}_i = \tilde{F}_Y^{-1}   Z_i $, it is not hard to see that $\tilde{Z}_1, \tilde{Z}_2, \dots , \tilde{Z}_b $ are i.i.d random variables with CDF $F_Z^{-1}   \tilde{F}_Y$,
since
\begin{equation}
	F_Z^{-1}   \tilde{F}_Y(t) = \Pr(Z_i < \tilde{F}_Y(t))= \Pr(\tilde{F}_Y^{-1}   Z_i < t).    
\end{equation}
Accordingly, we have the $\tilde{Z}_i$ associated EDF
\begin{equation}
	\tilde{F}^{-1}_{Z,b}(t) = \frac{1}{b} \sum_{i=1}^b \mathbf{1}_{\{ \tilde{Z}_i \leq t\}} = \frac{1}{b} \sum_{i=1}^b \mathbf{1}_{\{ Z_i \leq \tilde{F}_Y(t) \}} = F^{-1}_{Z,b}   \tilde{F}_Y (t).   
\end{equation}
By applying the DKW-inequality it holds for all $\varepsilon\geq\sqrt{\tfrac{1}{2b}\ln2}$
\begin{equation}
	\Pr\Bigl(\| F^{-1}_{Z,b}   \tilde{F}_Y(t) - F^{-1}_{Z}   \tilde{F}_Y (t)\|_{\infty} > \varepsilon \Bigr) \le e^{-2b\varepsilon^2}.
\end{equation}
Since $F_Z^{-1}$ is continuous with a compact convex domain $[0,1]$, $F_Z^{-1}$ satisfies Lipschitz continuity with a Lipschitz constant $\delta$ and it holds
\begin{equation}
	\begin{aligned}
		& \|  F^{-1}_{Z,b}   \tilde{F}_Y(t) - F^{-1}_{Z}   F_Y (t)\|_{\infty}   \\ 
		& \leq \|  F^{-1}_{Z,b}   \tilde{F}_Y(t) - F^{-1}_{Z}   \tilde{F}_Y (t)\|_{\infty} 
		+ \|  F^{-1}_{Z}   \tilde{F}_Y(t) - F^{-1}_{Z}   F_Y (t)\|_{\infty} \\
		& \leq \|  F^{-1}_{Z,b}   \tilde{F}_Y(t) - F^{-1}_{Z}   \tilde{F}_Y (t)\|_{\infty} + \delta \gamma,
	\end{aligned}    
\end{equation}
then for $\omega \in  \{ \|F^{-1}_{Z,b}   \tilde{F}_Y - F^{-1}_{Z}   F_Y\|_{\infty} > \varepsilon \}$ we have 
\begin{equation}
	\omega \in  \{ \| F^{-1}_{Z,b}   \tilde{F}_Y - F^{-1}_{Z}   \tilde{F}_Y\|_{\infty} > \varepsilon - \delta \gamma \}.
\end{equation}
Therefore, for all $ \varepsilon- \delta \gamma \geq\sqrt{\tfrac{1}{2b}\ln2}$ we have 
\begin{equation}
	\Pr\Bigl(\| F^{-1}_{Z,b}   \tilde{F}_Y (t) - F^{-1}_{Z}   F_Y (t)\|_{\infty} > \varepsilon \Bigr) \le e^{-2b(\varepsilon- \delta \gamma)^2}. \qed 
\end{equation} 

\newpage

\subsection*{B \quad Complementary study of $k, k'$ }
\begin{figure}[h!]
	\center
	\begin{tabular}{l|lll}
		\toprule
		\slashbox{$k$}{$k'$}& 0 & 0.1 & 1 \\ \midrule
		$10^{-4}$ & 14.8 & 15.2 & 14.9 \\
		$10^{-3}$ & \textbf{13.0} &15.0 & 15.5 \\
		0.1 & 20.9 & 17.9 & 16.7 \\
		1   & 22.0 & 22.2 & 21.5 \\
		10  & 22.5 & 19.3 & 22.7 \\
		\bottomrule
	\end{tabular}
	\vspace{0.2cm}
	\caption{Complementary FID scores to Fig. 3 (Left h).}
\end{figure}

\section*{C \quad  More Hyperparameter Studies}

\begin{figure}[h!]
	\centering
	\begin{tabular}{c}
		\includegraphics[width=1\linewidth]{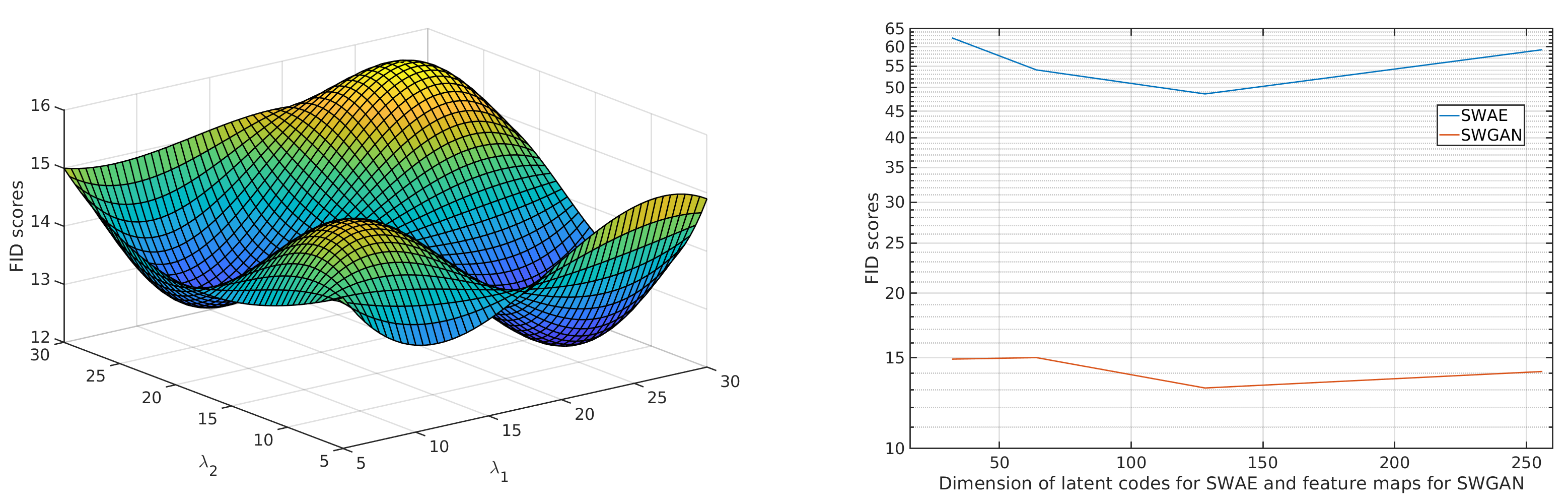}
	\end{tabular}
	\caption{The FID scores of various $\lambda_1, \lambda_2$ (left) and dimensions of latent codes for SWAE and feature maps for SWGAN (right). We can see that the FID scores are not very sensitive to the changes of $\lambda_1, \lambda_2$, while the optimal dimension for SWAE and SWGAN is $128$.}
	
	\label{fig:sotblock}
	\vspace{-0.5cm}
\end{figure}

\newpage
\subsection*{D \quad  Model Stability Study}

\begin{table}[h!]
	\centering
	\resizebox{0.85\textwidth}{!}{
		\begin{tabular}{ccccc}
			\toprule
			& ResNet (w/ norm) & ResNet (w/o norm) & ConvNet (w/ norm) & ConvNet (w/o norm) \\ \hline
			CTGAN  & 16.0       &16.5	&19.5	&19.7            \\ 
			& 
			\includegraphics[width=0.2\linewidth]{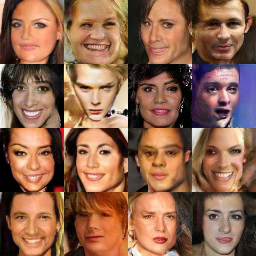}&
			\includegraphics[width=0.2\linewidth]{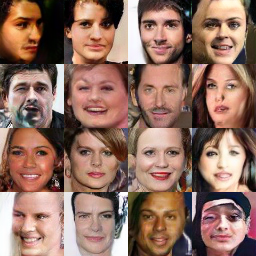}&
			\includegraphics[width=0.2\linewidth]{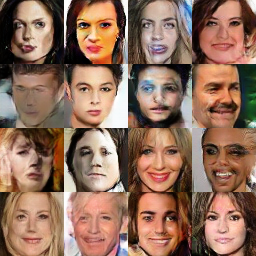}&
			\includegraphics[width=0.2\linewidth]{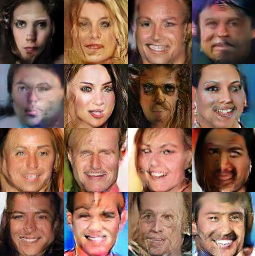}\\ \hline

			SWG & 24.3 & 29.1 & 22.2 &28.5 \\ 
			
			& \includegraphics[width=0.2\linewidth]{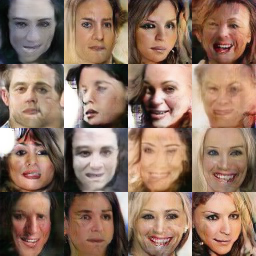}&
			\includegraphics[width=0.2\linewidth]{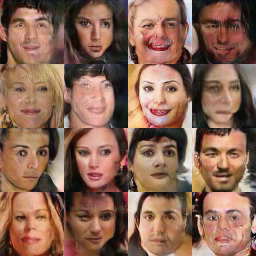}&
			\includegraphics[width=0.2\linewidth]{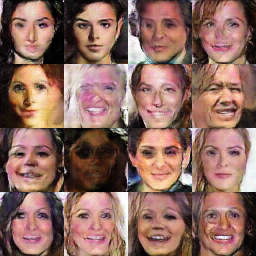}&
			\includegraphics[width=0.2\linewidth]{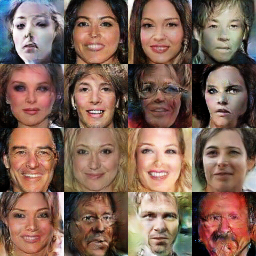}\\ \hline

			\textbf{SWGAN} & \textbf{13.0}  & \textbf{14.8}  & \textbf{19.2} & \textbf{18.8}  \\ 
			
			& \includegraphics[width=0.2\linewidth]{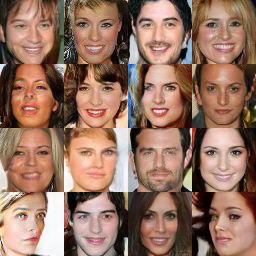}&
			\includegraphics[width=0.2\linewidth]{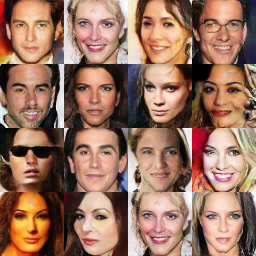}&
			\includegraphics[width=0.2\linewidth]{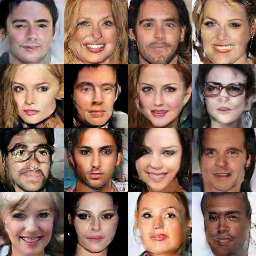}&
			\includegraphics[width=0.2\linewidth]{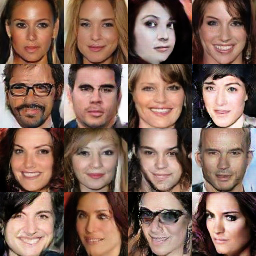}\\ \bottomrule
			
			WAE-MMD &64.0   &61.8 &55.8 &67.8 \\ 
			& 
			\includegraphics[width=0.2\linewidth]{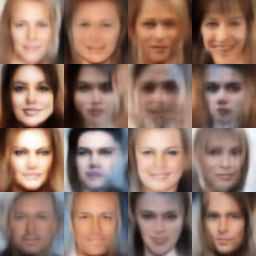}&
			\includegraphics[width=0.2\linewidth]{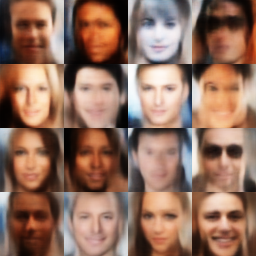}&
			\includegraphics[width=0.2\linewidth]{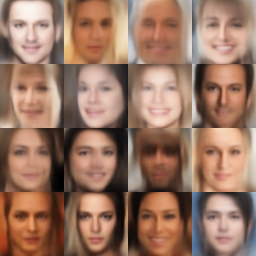}&
			\includegraphics[width=0.2\linewidth]{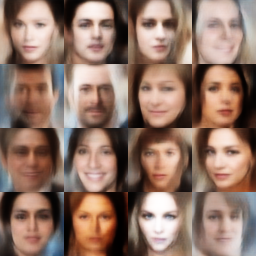}\\ \bottomrule

			AAE (WAE-GAN)& \textbf{62.3}  &\textbf{56.7}	& \textbf{48.3} &66.1	 \\ 
			& 
			\includegraphics[width=0.2\linewidth]{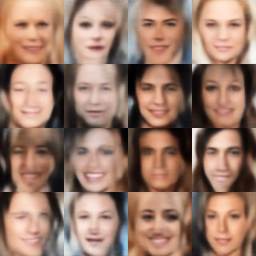}&
			\includegraphics[width=0.2\linewidth]{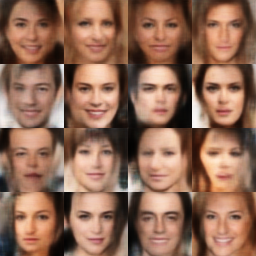}&
			\includegraphics[width=0.2\linewidth]{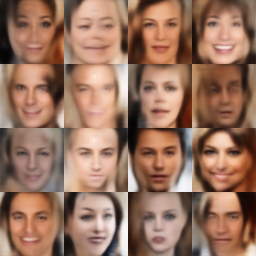}&
			\includegraphics[width=0.2\linewidth]{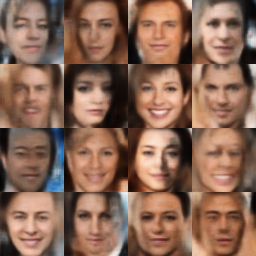}\\ \bottomrule

			\textbf{SWAE} & 63.2   & 59.1  &65.2 & \textbf{48.6}           \\ 
			& 
			\includegraphics[width=0.2\linewidth]{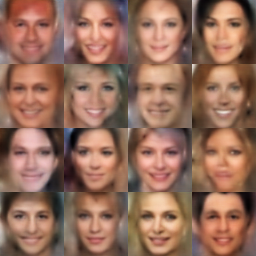}&
			\includegraphics[width=0.2\linewidth]{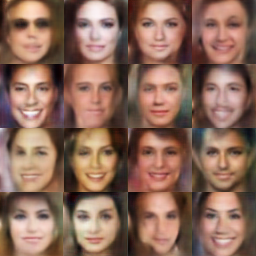}&
			\includegraphics[width=0.2\linewidth]{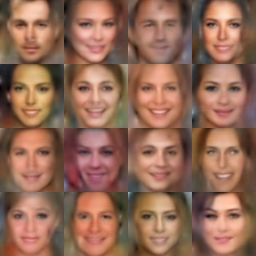}&
			\includegraphics[width=0.2\linewidth]{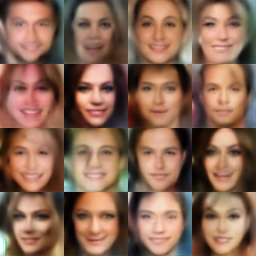}\\ \bottomrule

		\end{tabular}
	}
	\caption{FID scores of various architectures on CelebA. 
		The optimal architectures are ConvNet for SWG, WAE-MMD, AAE (WAE-GAN), ResNet for CTGAN, SWGAN, and ConvNet without normalization (w/o norm) for SWAE.}
\end{table}

\newpage
\subsection*{E \quad  Architecture Details of Our Proposed Models}
\begin{table}[h!]
	\centering
	\small
	\begin{tabular}{cccc}
		\toprule
		
		\textbf{Encoder}        &  Kernel size          & Resampling           & Output shape           \\\midrule          
		NearestNeighbor  & --   & Down           & $3\times16\times16$           \\
		Linear   & --    & --           & 128           \\
		3 Primal SWD blocks  & --    & --           & 128           \\
		\bottomrule
		
		\textbf{Decoder}            &          &         &            \\ \midrule
		Noise         & --            & --            & 128           \\
		Linear          & --            & --            & $64\times8\times8$            \\
		2 (Conv, ELU) blocks  & $3\times3$    & --           & $64\times8\times8$           \\
		NearestNeighbor  & --   & Up           & $64\times16\times16$           \\
		2 (Conv, ELU) blocks  & $3\times3$    & --           & $64\times16\times16$             \\
		NearestNeighbor  & --   & Up           & $64\times32\times32$           \\
		2 (Conv, ELU) blocks  & $3\times3$    & --           & $64\times32\times32$ \\  
		NearestNeighbor  & --   & Up           & $64\times64\times64$           \\
		2 (Conv, ELU) blocks  & $3\times3$    & --           & $64\times64\times64$ \\  
		Conv, tanh  & $3\times3$    & --           & $3\times64\times64$           \\
		\bottomrule
	\end{tabular}
	\vspace{0.2cm}
	\caption{Network architecture for SWAE}
	\label{tab:nwae}
	\vspace{0.5cm}
	\begin{tabular}{cccc}
		\toprule
		\textbf{Generator}            &  Kernel size      & Resampling     & Output shape           \\ \midrule
		Noise         & --            & --            & 128            \\
		Linear          & --            & --            & $128\times4\times4$            \\
		2 (Res, ReLU) blocks  & $3\times3$    & Up           & $128\times8\times8$           \\
		2 (Res, ReLU) blocks  & $3\times3$    & Up           & $128\times16\times16$           \\
		2 (Res, ReLU) blocks  & $3\times3$    & Up           & $128\times32\times32$           \\
		2 (Res, ReLU) blocks  & $3\times3$    & Up           & $128\times64\times64$           \\
		Conv, tanh  & $3\times3$    & --           & $3\times64\times64$           \\
		\bottomrule

		\textbf{Discriminator}            &          &         &            \\ \midrule
		2 (Res, ReLU) blocks  & $3\times3$    & Down           & $128\times32\times32$           \\
		2 (Res, ReLU) blocks  & $3\times3$    & Down           & $128\times16\times16$           \\
		2 (Res, ReLU) blocks  & $3\times3$    & Down           & $128\times8\times8$           \\
		2 (Res, ReLU) blocks  & $3\times3$    & --           & $128\times8\times8$           \\
		Linear   & --    & --           & 128    \\
		4 (Dual, LeakyReLU) SWD blocks  & --    & --           & 128    \\
		\bottomrule
	\end{tabular}
	\caption{Network architecture for SWGAN}
	\label{tab:netswgan}
\end{table}

\newpage
\section*{F \quad  More Visual Results for SWAE, SWGAN and Compared Methods}
\begin{table*}[!h]
	\tiny
	\centering
	\resizebox{\linewidth}{!}{%
		\begin{tabular}{cccc}
			\toprule
			VAE & WAE-MMD & AAE (WAE-GAN) & \textbf{SWAE}\\ \hline
			\includegraphics[width=0.15\linewidth]{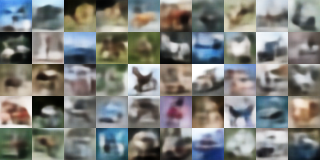}&
			\includegraphics[width=0.15\linewidth]{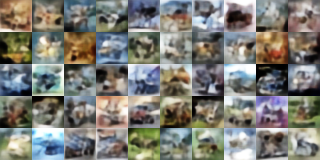}&
			\includegraphics[width=0.15\linewidth]{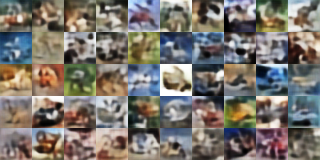}&
			\includegraphics[width=0.15\linewidth]{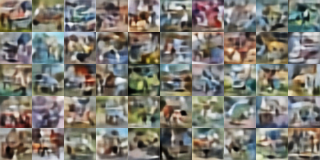}\\
			
			\includegraphics[width=0.15\linewidth]{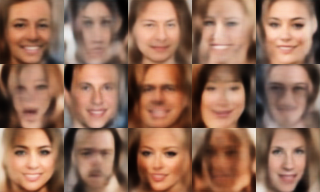}&
			\includegraphics[width=0.15\linewidth]{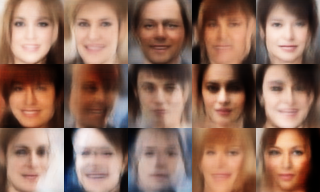}&
			\includegraphics[width=0.15\linewidth]{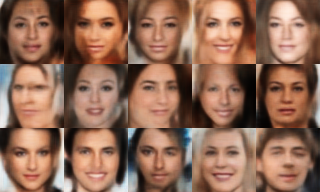}&
			\includegraphics[width=0.15\linewidth]{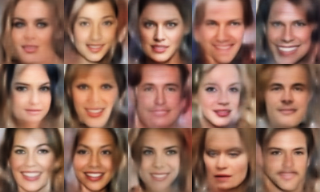}\\
			
			\midrule
			SWG & CTGAN & WGAN-GP & \textbf{SWGAN}\\ \hline
			\includegraphics[width=0.15\linewidth]{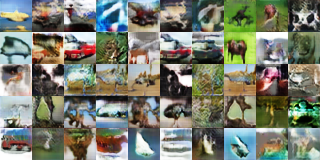}&
			\includegraphics[width=0.15\linewidth]{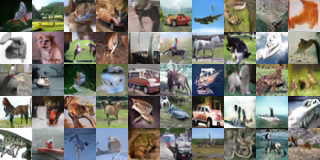}&
			\includegraphics[width=0.15\linewidth]{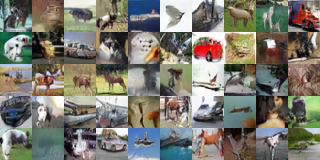}&
			\includegraphics[width=0.15\linewidth]{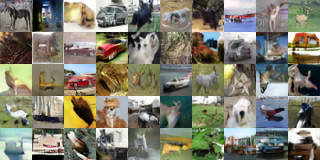}\\ 
			
			\includegraphics[width=0.15\linewidth]{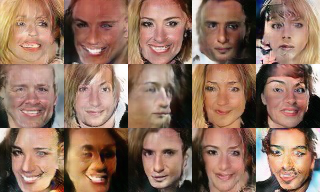}&
			\includegraphics[width=0.15\linewidth]{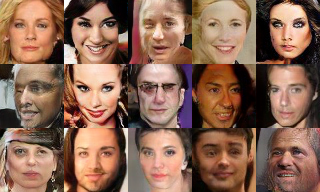}&
			\includegraphics[width=0.15\linewidth]{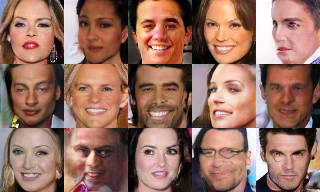}&
			\includegraphics[width=0.15\linewidth]{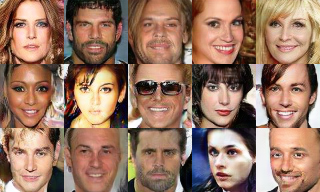}\\

			\includegraphics[width=0.15\linewidth]{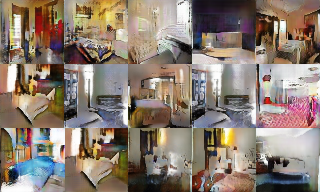}&
			\includegraphics[width=0.15\linewidth]{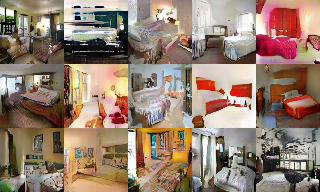}&
			\includegraphics[width=0.15\linewidth]{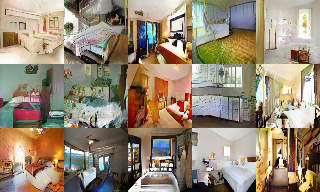}&
			\includegraphics[width=0.15\linewidth]{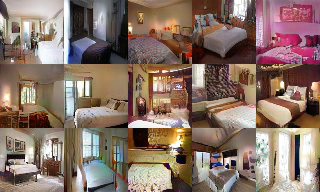}\\
			\bottomrule
		\end{tabular}
	}
	\caption{Visual results of AE-based (top 2 rows) and GAN (bottom 3 rows) models on CIFAR-10, CelebA and LSUN.}
	\vspace{0.5cm}
	\label{fig:vae_gan_results_supp}
	\resizebox{\linewidth}{!}{%
		\begin{tabular}{cc}
			\includegraphics[width=0.4\linewidth, height=3cm]{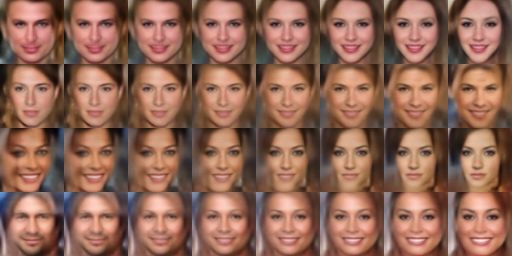}&
			\includegraphics[width=0.4\linewidth, height=3cm]{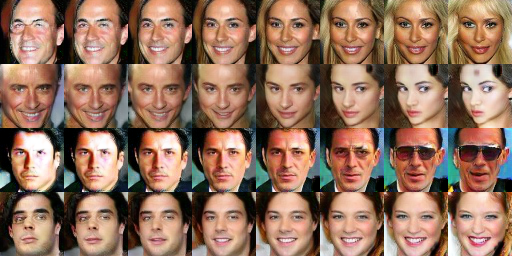}\\
		\end{tabular}}
		\caption{Interpolation results of the proposed SWAE (left) and SWGAN (right) models on CelebA.}
		\label{tab:inter_celeba}
	\end{table*}

	\newpage
	\section*{G \quad  More Visual Results for PG-WGAN and PG-SWGAN}
	\begin{figure}[h!]
		
		\centering
		\resizebox{\linewidth}{!}{%
			\small
			\begin{tabular}{cc}
				\toprule
				
				PG-WGAN & \textbf{PG-SWGAN}\\ \hline
				
				\includegraphics[width=0.45\linewidth]{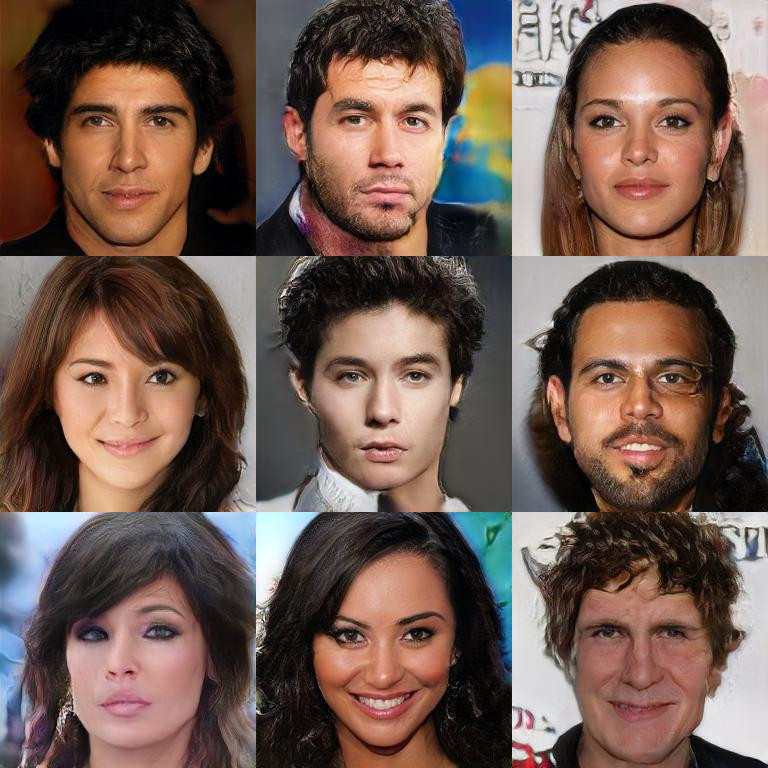} &
				\includegraphics[width=0.45\linewidth]{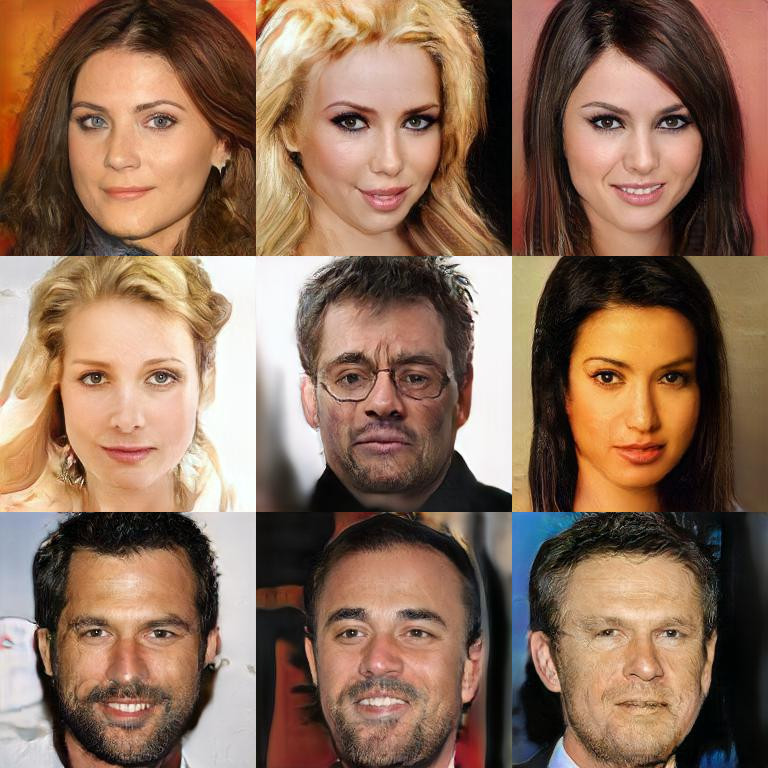}\\
				\includegraphics[width=0.45\linewidth]{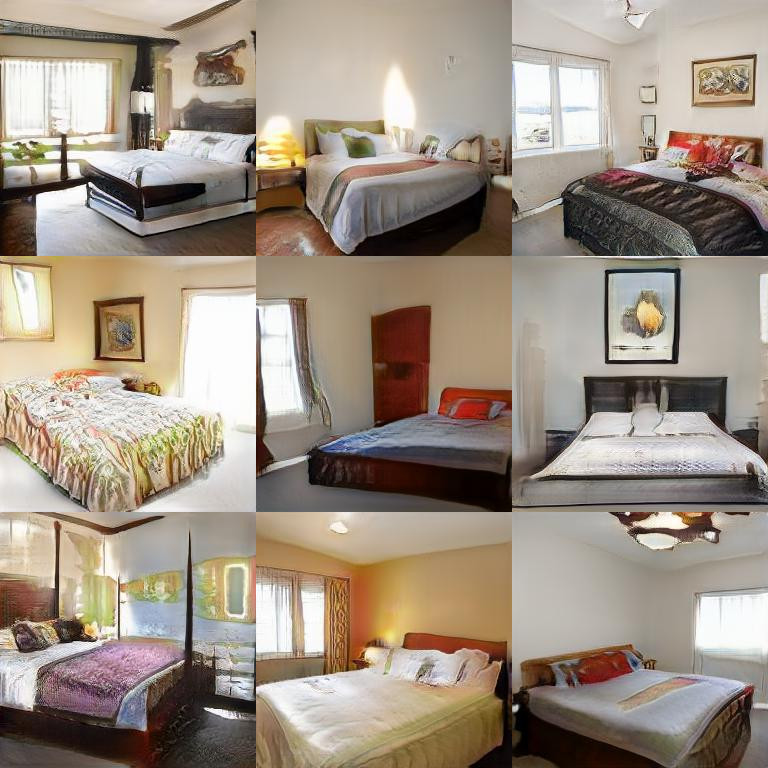}&
				\includegraphics[width=0.45\linewidth]{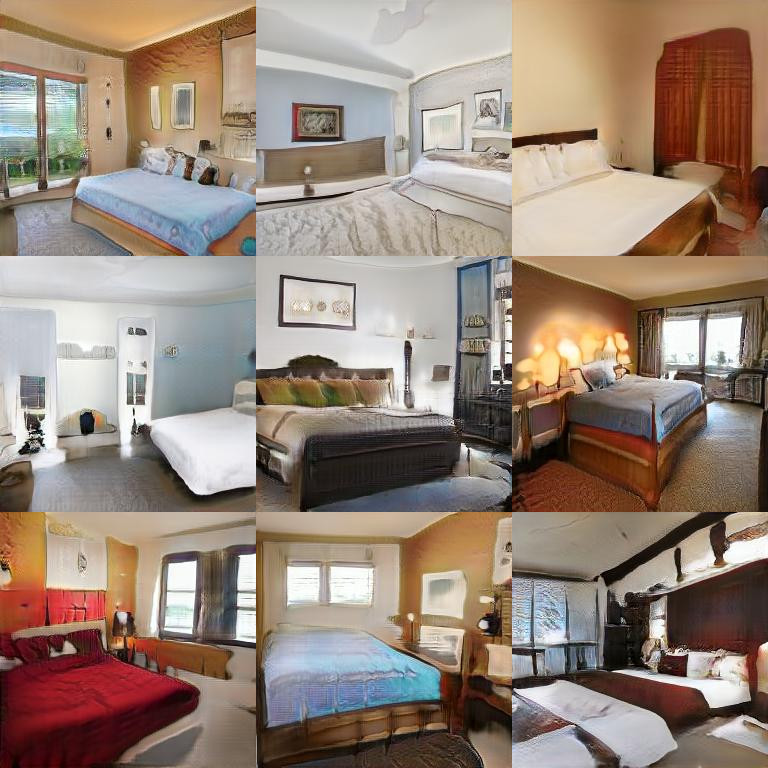}\\
				\bottomrule
			\end{tabular}
		}
		\caption{Visual results for PG-WGAN and PG-SWGAN on CelebA-HQ and LSUN.}
		\label{fig:pgtraining_visual}
	\end{figure}

\end{document}